\documentclass{article}
\usepackage[dvipsnames]{xcolor}

     \usepackage[preprint,nonatbib]{neurips_2020}

\usepackage[utf8]{inputenc} 
\usepackage[T1]{fontenc}    
\usepackage{hyperref}       
\usepackage{url}            
\usepackage{booktabs}       
\usepackage{amsfonts}       
\usepackage{nicefrac}       
\usepackage{microtype}      
\usepackage{graphicx}
\usepackage{amssymb}
\usepackage{amsmath}
\usepackage{amsthm}
\usepackage{hyperref}
\usepackage{tabularx}

\usepackage{tabularx,ragged2e}
\usepackage{multirow}
\usepackage{threeparttable}
\usepackage{microtype}
\usepackage{wrapfig}

\newcommand{\pathi}{\mathbf{l}}

\title{Path Integral Based Convolution and Pooling for Graph Neural Networks}

\author{%
  Zheng Ma\\
  Department of Physics\\
  Princeton University\\
  \texttt{zhengm@princeton.edu} \\
  \And
   Junyu Xuan \\
   Centre for Artificial Intelligence\\
   Faculty of Engineering and Information Technology\\
   University of Technology Sydney \\
   \texttt{junyu.xuan@uts.edu.au} \\
   \And
   Yu Guang Wang \\
   Max Planck Institute for Mathematics in the Sciences\\
   \& School of Mathematics and Statistics\\
   University of New South Wales\\
   \texttt{yuguang.wang@mis.mpg.de} \\
  \And
   Ming Li \\
   Department of Educational Technology\\
  Zhejiang Normal University \\
   \texttt{mingli@zjnu.edu.cn} \\
   \And
   Pietro Li\`{o} \\
   Department of Computer Science and Technology \\
   University of Cambridge\\
   \texttt{Pietro.Lio@cl.cam.ac.uk} \\
}

\begin{document}

\maketitle

\begin{abstract}
Graph neural networks (GNNs) extends the functionality of traditional neural networks to graph-structured data. Similar to CNNs, an optimized design of graph convolution and pooling is key to success. Borrowing ideas from physics, we propose a path integral based graph neural networks (PAN) for classification and regression tasks on graphs. Specifically, we consider a convolution operation that involves every path linking the message sender and receiver with learnable weights depending on the path length, which corresponds to the maximal entropy random walk. It generalizes the graph Laplacian to a new transition matrix we call \emph{maximal entropy transition} (MET) matrix derived from a path integral formalism. Importantly, the diagonal entries of the MET matrix are directly related to the subgraph centrality, thus providing a natural and adaptive pooling mechanism. PAN provides a versatile framework that can be tailored for different graph data with varying sizes and structures. We can view most existing GNN architectures as special cases of PAN. Experimental results show that PAN achieves state-of-the-art performance on various graph classification/regression tasks, including a new benchmark dataset from statistical mechanics we propose to boost applications of GNN in physical sciences. 
\end{abstract}

\section{Introduction}
The triumph of convolutional neural networks (CNNs) has motivated researchers to develop similar architectures for graph-structured data. The task is challenging due to the absence of regular grids. One notable proposal is to define convolutions in the Fourier space \cite{BrZaSzLe2013,Bronstein_etal2017}. This method relies on finding the spectrum of the graph Laplacian $I-D^{-1}A$ or $I-D^{-\frac{1}{2}}AD^{-\frac{1}{2}}$ and then applies filters to the components of input signal $X$ under the corresponding basis, where $A$ is the adjacency matrix of the graph, and $D$ is the corresponding degree matrix. Due to the high computational complexity of diagonalizing the graph Laplacian, people have proposed many simplifications \cite{defferrard2016convolutional, KiWe2017}.

The graph Laplacian based methods essentially rely on message passing \cite{gilmer2017neural} between directly connected nodes with equal weights shared among all edges, which is at heart a generic random walk (GRW) defined on graphs. It can be seen most obviously from the GCN model \cite{KiWe2017}, where the normalized adjacency matrix is directly applied to the left-hand side of the input. In statistical physics, $D^{-1}A$ is known as the transition matrix of a particle doing a random walk on the graph, where the particle hops to all directly connected nodes with equiprobability. Many direct space-based methods \cite{node2vec_2016,LiTaBrZe2015,velivckovic2017graph,Planetoid_2016} can be viewed as generalizations of GRW, but with biased weights among the neighbors.

In this paper, we go beyond the GRW picture, where information necessarily dilutes when a path branches, and instead consider every path linking the message sender and receiver as the elemental unit in message passing. Inspired by the path integral formulation developed by Feynman \cite{feynman2010quantum,feynman1948space}, we propose a graph convolution that assigns trainable weights to each path depending on its length. This formulation results in a \emph{maximal entropy transition} (MET) matrix, which is the counterpart of graph Laplacian in GRW. By introducing a fictitious temperature, we can continuously tune our model from a fully localized one (MLP) to a spectrum based model. Importantly, the diagonal of the MET matrix is intimately related to the subgraph centrality, and thus provides a natural pooling method without extra computations. We call this complete path integral based graph neural network framework PAN. 

We demonstrate that PAN outperforms many popular architectures on benchmark datasets. We also introduce a new dataset from statistical mechanics, which overcomes the lack of explanability and tunability of many previous ones. The dataset can serve as another benchmark, especially for boosting applications of GNN in physical sciences. This dataset again confirms that PAN has a faster convergence rate, higher prediction accuracy, and better stability compared to many counterparts. 

\section{Path Integral Based Graph Convolution}
\paragraph{Path integral and MET matrix} Feynman's path integral formulation \cite{feynman2010quantum,Zinn-Justin:2009} interprets the probability amplitude $\phi(x,t)$ as a weighted average in the configuration space, where the contribution from $\phi_0(x)$ is computed by summing over the influences (denoted by $e^{iS[\mathbf{x},\mathbf{\Dot{x}}]}$) from all paths connecting itself and $\phi(x,t)$. This formulation has been later extensively used in statistical mechanics and stochastic processes \cite{kleinert2009path}. We note that this formulation essentially constructs a convolution by considering the contribution from all possible paths in the continuous space. 
\begin{figure}[th]
\vskip -4mm
\centering
\begin{minipage}{0.8\textwidth}
    \includegraphics[width=\textwidth]{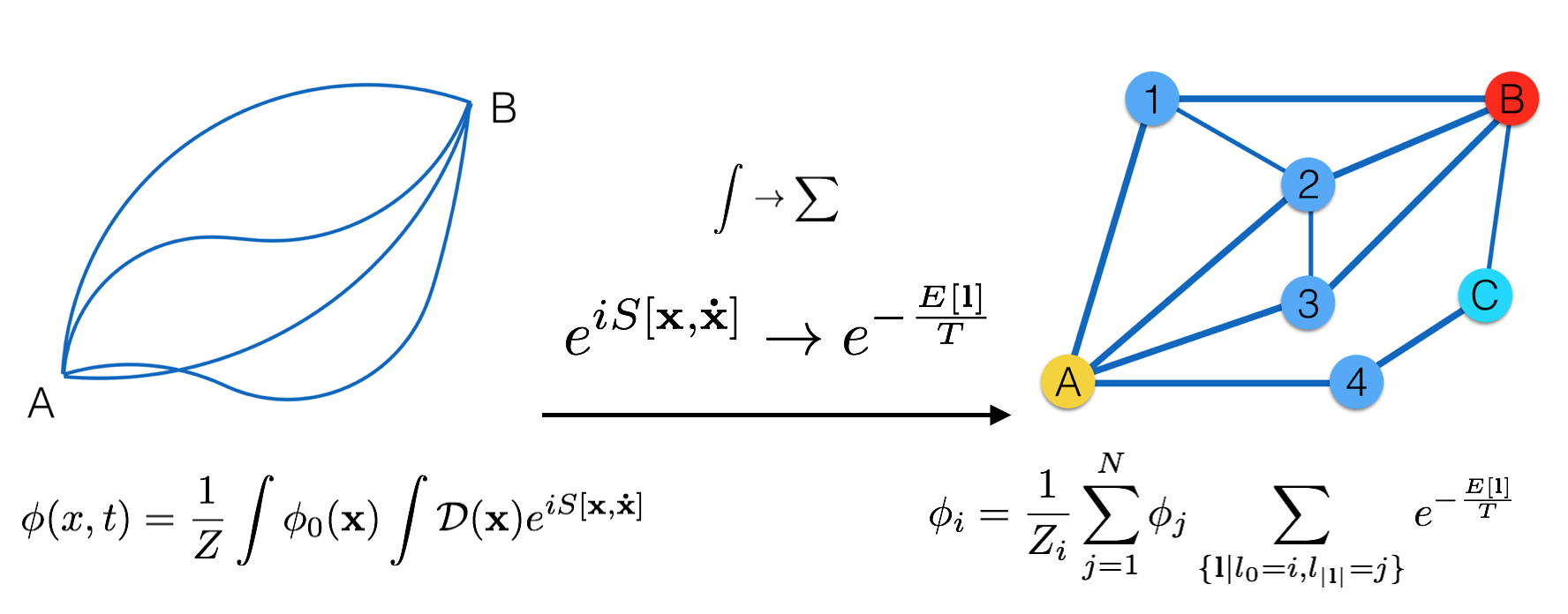}
\end{minipage}
\vspace{-2mm}
\caption{A schematic analogy between the original path integral formulation in continuous space (left) and the discrete version for a graph (right). Symbols are defined in the text.}
\label{fig:panconv}
\vskip -0.1in
\end{figure}
Using this idea, but modified for discrete graph structures, we can heuristically propose a statistical mechanics model on how information is shared between different nodes on a given graph.
In the most general form, we write observable $\phi_i$ at the $i$-th node for a graph with $N$ nodes as
\begin{equation} \label{eq:stat}
\phi_i=\frac{1}{Z_i}\sum_{j=1}^{N}\phi_j \sum_{\{\mathbf l|l_0=i,l_{|\pathi|}=j\}}e^{-\frac{E[\mathbf l]}{T}},
\end{equation}
where $Z_i$ is the normalization factor known as the \textit{partition function} for the $i$-th node. Here a path $\pathi$ is a sequence of connected nodes $(l_0l_1\dots l_{|\pathi|})$ where $A_{l_il_{i+1}}=1$, and the length of the path is denoted by $|\pathi|$. In Figure~\ref{fig:panconv} we draw the analogy between our discrete version and the original formulation. It is straightforward to see that the integral should now be replaced by a summation, and $\phi_0(x)$ only resides on nodes. Since a statistical mechanics perspective is more proper in our case, we directly change the exponential term, which is originally an integral of Lagrangian, to a Boltzmann's factor with fictitious energy $E[\pathi]$ and temperature $T$; we choose Boltzmann's constant $k_B=1$. Nevertheless, we still exploit the fact that the energy is a functional of the path, which gives us a way to weight the influence of other nodes through a certain path. The fictitious temperature controls the excitation level of the system, which reflects that to what extent information is localized or extended. In practice, there is no need to learn the fictitious temperature or energy separately, instead the neural networks can directly learn the overall weights, as will be made clearer later.

To obtain an explicit form of our model, we now introduce some mild assumptions and simplifications. Intuitively, we know that information quality usually decays as the path between the message sender and the receiver becomes longer, thus it is reasonable to assume that the energy is not only a functional of path, but can be further simplified as a function that solely depends on the length of the path. In the random walk picture, this means that the hopping is equiprobable among all the paths that have the same length, which maximizes the Shannon entropy of the probability distribution of paths globally, and thus the random walk is given the name maximal entropy random walk \cite{burda2009localization} \footnote{For a weighted graph, a feasible choice for the functional form of the energy could be $E(l_{\rm eff})$, where the effective length of the path $l_{\rm eff}$ can be defined as a summation of the inverse of weights along the path, i.e.  $l_{\rm eff}=\sum_{i=0}^{|l|-1}1/w_{l_il_{i+1}}$.}. By first conditioning on the length of the path, we can introduce the overall $n$-th layer weight $k(n;i)$ for node $i$ by
\begin{equation} \label{eq:kn}
k(n;i)=\frac{1}{Z_i}{\sum_{j=1}^{N}g(i,j;n)}e^{-\frac{E(n)}{T}},
\end{equation}
where $g(i,j;n)$ denotes the number of paths between nodes $i$ and $j$ with length of $n$, or \textit{density of states} for the energy level $E(n)$ with respect to nodes $i$ and $j$, and the summation is taken over all nodes of the graph. Intuitively, node $j$ with larger $g(i,j;n)$ means that it has more channels to talk with node $i$, thus may impose a greater influence on node $i$ as the case in our formulation. For example, in Figure~\ref{fig:panconv}, nodes $B$ and $C$ are both two-step away from $A$, but $B$ has more paths connecting $A$ and would be assigned with a larger weight as a consequence. Presumably, the energy $E(n)$ is an increasing function of $n$, which leads to a decaying weight as $n$ increases.\footnote{This does not mean that $k(n;i)$ must necessarily be a decreasing function, as $g(i,j;n)$ grows exponentially in general. It would be valid to apply a cutoff as long as $E(n)\gg nT\ln \lambda_1$ for large $n$, where $\lambda_1$ is the largest eigenvalue of the adjacency matrix $A$.} By applying a cutoff of the maximal path length $L$, we exchange the summation order in \eqref{eq:stat} to obtain
\begin{equation}
\phi_i=\sum_{n=0}^{L}k(n;i)\sum_{j=1}^{N}\frac{g(i,j;n)}{\sum_{s=1}^{N}g(i,s;n)}\phi_j
=\frac{1}{Z_i}\sum_{n=0}^{L}e^{-\frac{E(n)}{T}}\sum_{j=1}^{N}g(i,j;n)\phi_j,
\label{eq:sumkn}
\end{equation}
where the partition function can be explicitly written as
\begin{equation} \label{eq:partition}
Z_i=\sum_{n=0}^{L}e^{-\frac{E(n)}{T}}\sum_{j=1}^{N}g(i,j;n).
\end{equation}
A nice property of this formalism is that we can easily compute $g(i,j;n)$ by raising the power of the adjacency matrix $A$ to $n$, which is a well-known property of the adjacency matrix from graph theory, i.e., $g(i,j;n)=A^n_{ij}$. 
Plug in \eqref{eq:sumkn} we now have a group of self-consistent equations governed by a transition matrix $M$ (a counterpart of the \textit{propagator} in quantum mechanics), which can be written in the following compact form 
\begin{equation} \label{eq:Propagator2}
M=Z^{-1}\sum_{n=0}^{L}e^{-\frac{E(n)}{T}}A^n,
\end{equation}
where ${\rm diag}(Z)_i=Z_i$.
We call the matrix $M$ \emph{maximal entropy transition} (MET) matrix, with regard to the fact that it realizes maximal entropy under the microcanonical ensemble. This transition matrix replaces the role of the graph Laplacian under our framework. 

More generally, one can constrain the paths under consideration to, for example, shortest paths or self-avoiding paths. Consequentially, $g(i,j;n)$ will take more complicated forms and the matrix $A^n$ needs to be modified accordingly. In this paper, we focus on the simplest scenario and apply no constraints for the simplicity of the discussion. 
\paragraph{PAN convolution} The \emph{eigenstates}, or the basis of the system $\{\psi_i\}$ satisfy $M\psi_i=\lambda_i\psi_i$.
Similar to the basis formed by the graph Laplacian, one can define graph convolution based on the spectrum of MET matrix, which now has a distinct physical meaning. However, it is computationally impractical to diagonalize $M$ in every iteration as it is updated. To reduce the computational complexity, we apply the trick similar to GCN \cite{KiWe2017} by directly multiplying $M$ to the left hand side of the input and accompanying it by another weight matrix $W$ on the right-hand side. The convolutional layer is then reduced to a simple form
\begin{equation} \label{eq:conv}
X^{(h+1)}=M^{(h)}X^{(h)}W^{(h)},
\end{equation}
where $h$ refers to the layer number. 
Applying $M$ to the input $X$ is essentially a weighted average among neighbors of a given node, which leads to the question that if the normalization consistent with the path integral formulation works best in a data-driven context. It has been consistently shown experimentally that a symmetric normalization usually gives better results \cite{KiWe2017,LNet,MaLiWa2019}. This observation might have an intuitive explanation. Most generally, one can consider the normalization $Z^{-\theta_1}\cdot Z^{-\theta_2}$, where $\theta_1+\theta_2=1$. There are two extreme situations. When $\theta_1=1$ and $\theta_2=0$, it is called random-walk normalization and the model can be understood as ``receiver-controlled", in the sense that the node of interest performs an average among all the neighbors weighted by the number of channels that connect them. On the contrary, when $\theta_1=0$ and $\theta_2=1$, the model becomes ``sender-controlled", since the weight is determined by the fraction of the flow coming out from the sender that is directed to the receiver. Because of the fact that for an undirected graph, the exact interaction between connected nodes are unknown, as a compromise, the symmetric normalization can outperform both extremes, even it may not be the optimal.   
This consideration leads us to a final perfection step that changes the normalization $Z^{-1}$ in $M$ to the symmetric normalized version. The convolutional layer then becomes 
\begin{equation}\label{eq:conv_symm_norm}
X^{(h+1)}=M^{(h)}X^{(h)}W^{(h)}=Z^{-1/2}\sum_{n=0}^{L}e^{-\frac{E(n)}{T}}A^n Z^{-1/2}X^{(h)}W^{(h)}.
\end{equation}   
We shall call this graph convolution \emph{PANConv}.

The optimal cutoff $L$ of the series depends on the intrinsic properties of the graph, which is represented by temperature $T$. Incorporating more terms is analogous to having more particles excited to the higher energy level at a higher temperature. For instance, in \emph{low-temperature limit}, $L=0$, the model is reduced to the MLP model. In the \emph{high-temperature limit}, all factors $\exp(-E(n)/T)$ are effectively one, and the term with the largest power dominates the summation. We can see it by 
$A^n=\sum_{i=1}^{N}\lambda_i^n \psi_i\psi_i^T$,
where $\lambda_1,\dots,\lambda_N$ is sorted in a descending order. By the Perron-Frobenius theorem, we may only keep the leading order term with the unique largest eigenvalue $\lambda_1$ when $n\rightarrow \infty$. We then reach a prototype of the high temperature model $X^{(h+1)}=(I+\psi_1\psi_1^T)X^{(h)}W^{(h)}$. The most suitable choice of the cutoff $L$ reflects the intrinsic dynamics of the graph.

\section{Path Integral Based Graph Pooling}
For graph classification and regression tasks, another critical component is the pooling mechanism, which enables us to deal with graph input with variable sizes and structures. Here we show that the PAN framework provides a natural ranking of node importance based on the MET matrix, intimately related to the subgraph centrality. This pooling scheme, denoted by PANPool, requires no further work aside from the convolution and can discover the underlying local motif adaptively.  
\paragraph{MET matrix and subgraph centrality} 
Many different ways to rank the ``importance" of nodes in a graph have been proposed in the complex networks community. The most straightforward one is the degree centrality (DC), which counts the number of neighbors, other more sophisticated measures include, for example, betweenness centrality (BC) and eigenvector centrality (EC) \cite{newman2018networks}. Although these methods do give specific measures of the global importance of the nodes, they usually fail to pick up local patterns. However, from the way CNNs work on image classifications, we know that it is the \textit{locally} representative pixels that matter.

Estrada and Rodriguez-Velazquez \cite{estrada2005subgraph} have shown that subgraph centrality is superior to the methods mentioned above in detecting local graph motifs, which are crucial to the analysis of many social and biological networks. The subgraph centrality computes a weighted sum of the number of self-loops with different lengths. Mathematically, it simply writes as $\sum_{k=0}^{\infty}(A^k)_{ii}/k!$ for node $i$.  
Interestingly, one immediately sees that the resemblance of this expression and the diagonal elements of the MET matrix. The difference is easy to explain. The summation in the MET matrix is truncated at maximal length $L$, and the weights for different path length $e^{\frac{E(n)}{T}}$ is learnable. In contrast, the predetermined weight $1/k!$ is a convenient choice to ensure the convergence of the summation and an analytical form of the result, which writes $\sum_{j=1}^{N}v_j^2(i)e^{\lambda_j}$, where $v_j(i)$ is the $i$-th element of the orthonormal basis associated with the eigenvalue $\lambda_j$.

Now it becomes clear that the MET matrix plays the role of a path integral-based convolution. Its diagonal elements $M_{ii}$ also automatically provides a measure of the importance of node $i$, thus enabling a pooling mechanism by sorting $M_{ii}$. Importantly, this pooling method has three main merits compared to the subgraph centrality. First, we can exploit the readily-computed MET matrix, thus circumvent extra computations, especially the direct diagonalization of the adjacency matrix in the case of subgraph centrality. Second, the weights are data-driven rather than predetermined, which can effectively adapt to different inputs.
Furthermore, the MET matrix is normalized \footnote{Notice that unlike the case in convolutions, the normalization is symmetric or not does not matter here. Here we only care about the diagonal terms, and different normalization methods will give the same result.}, which adds weights on the \textit{local} importance of the nodes, and can potentially avoid clustering around ``hubs" that are commonly seen in real-world ``scale-free" networks \cite{barabasi2016network}. 

The PAN Pooling strategy has similar physical explanations as the PAN convolution. In the low-temperature limit, for example, if we set the cut-off at $L=2$, the rank of $\sum_{n=0}^{L}e^{\frac{E(n)}{T}}A_{ii}^n$ is of the same order as the rank of degrees, and thus we recover the degree centrality. In the high-temperature limit, as $n \rightarrow \infty$, the sum is dominated by the magnitude of the $i$-th element of the orthonormal basis associated with the largest eigenvalue of $A$, thus the corresponding ranking is reduced to the ranking of the eigenvector centrality. By tuning $L$, PANPool provides a flexible strategy that can adapt to the ``sweet spot" of the input. 

\begin{figure}[th]
    \centering
     \begin{minipage}{\textwidth}
     \centering	
     \begin{minipage}{0.19\textwidth}
     \includegraphics[width=\textwidth]{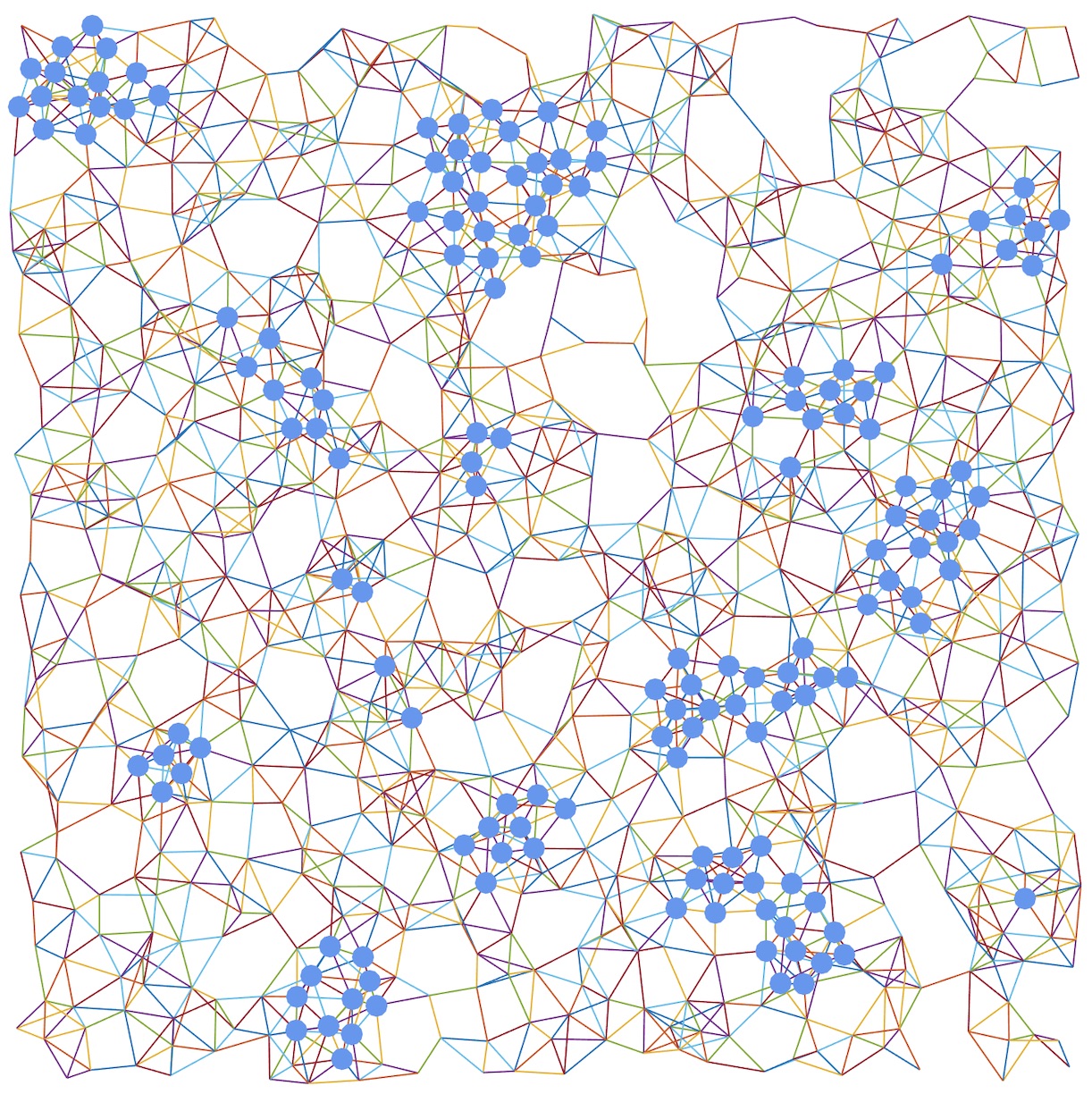}
     \end{minipage}
     \begin{minipage}{0.19\textwidth}
     \includegraphics[width=\textwidth]{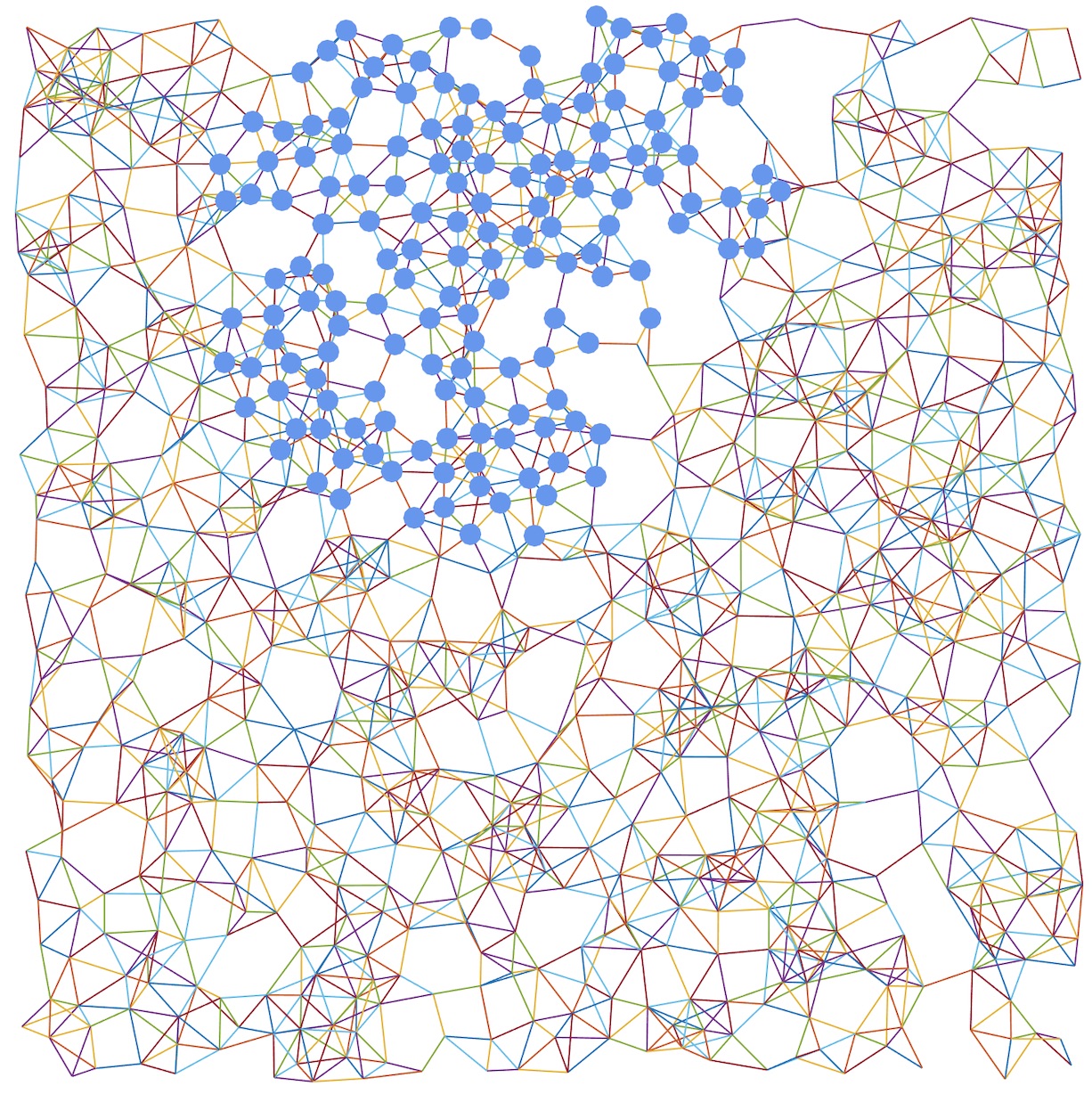}
     \end{minipage}
     \begin{minipage}{0.19\textwidth}
     \includegraphics[width=\textwidth]{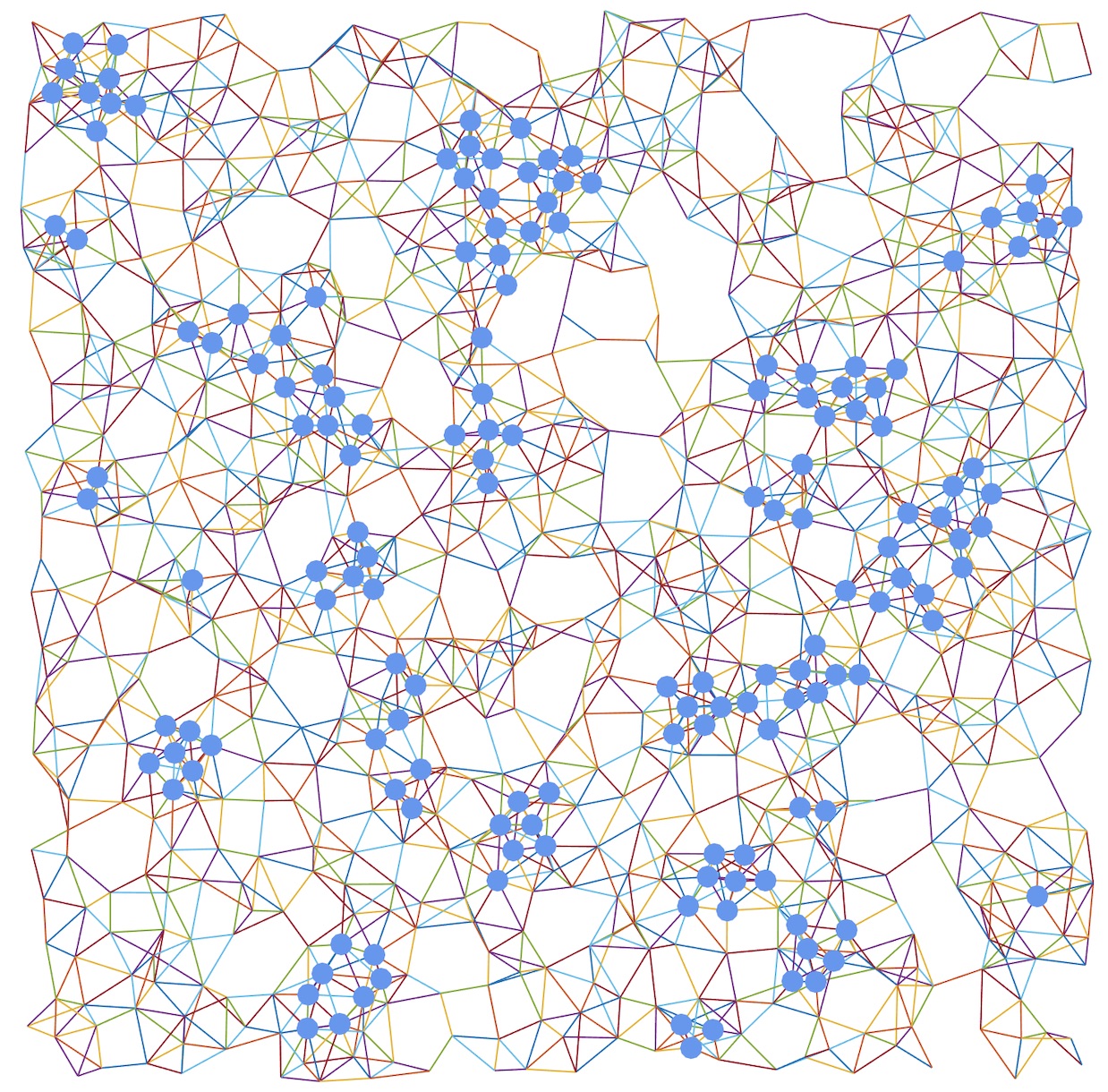}
     \end{minipage}
     \begin{minipage}{0.19\textwidth}
     \includegraphics[width=\textwidth]{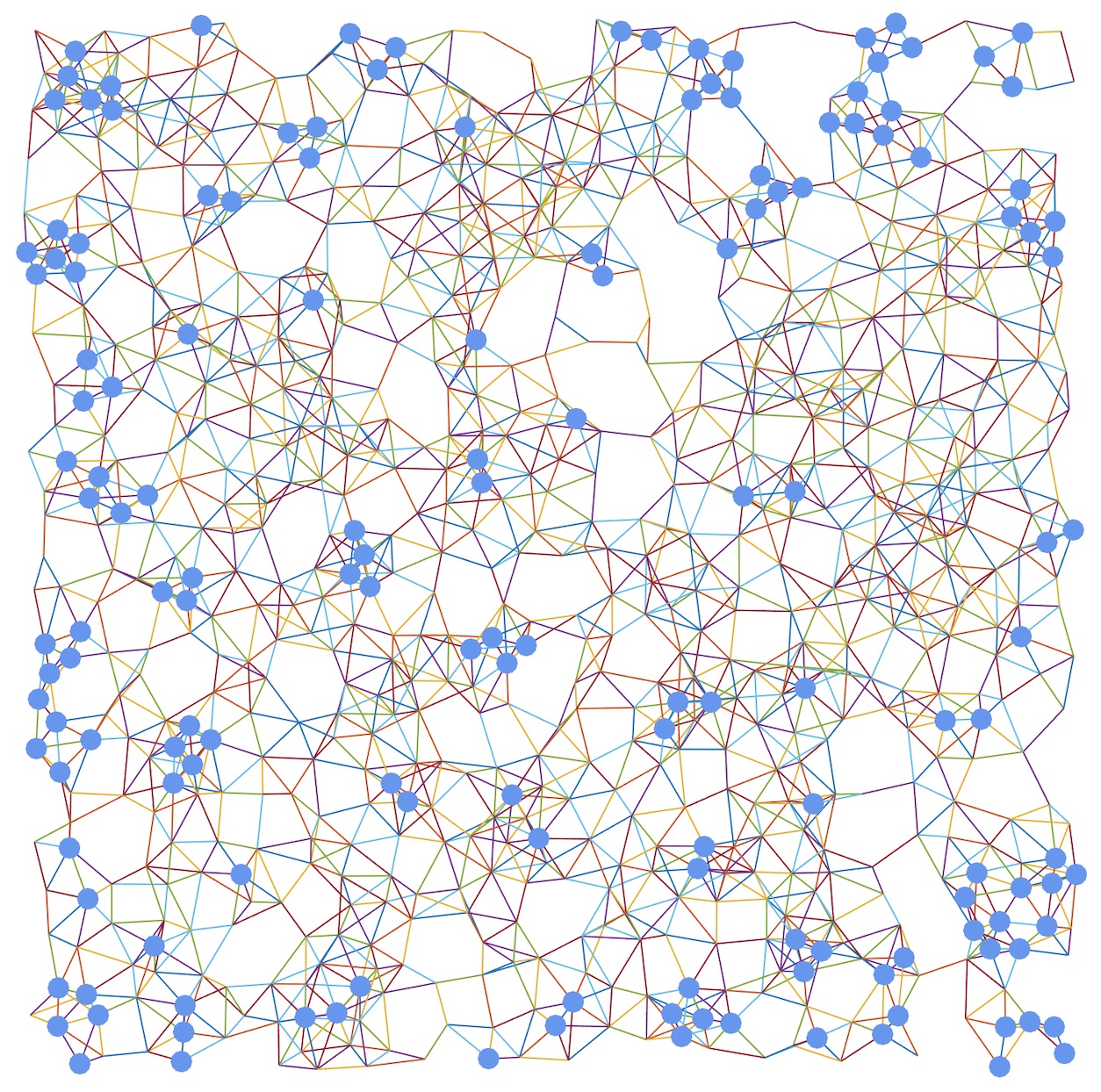}
     \end{minipage}
     \begin{minipage}{0.19\textwidth}
     \includegraphics[width=\textwidth]{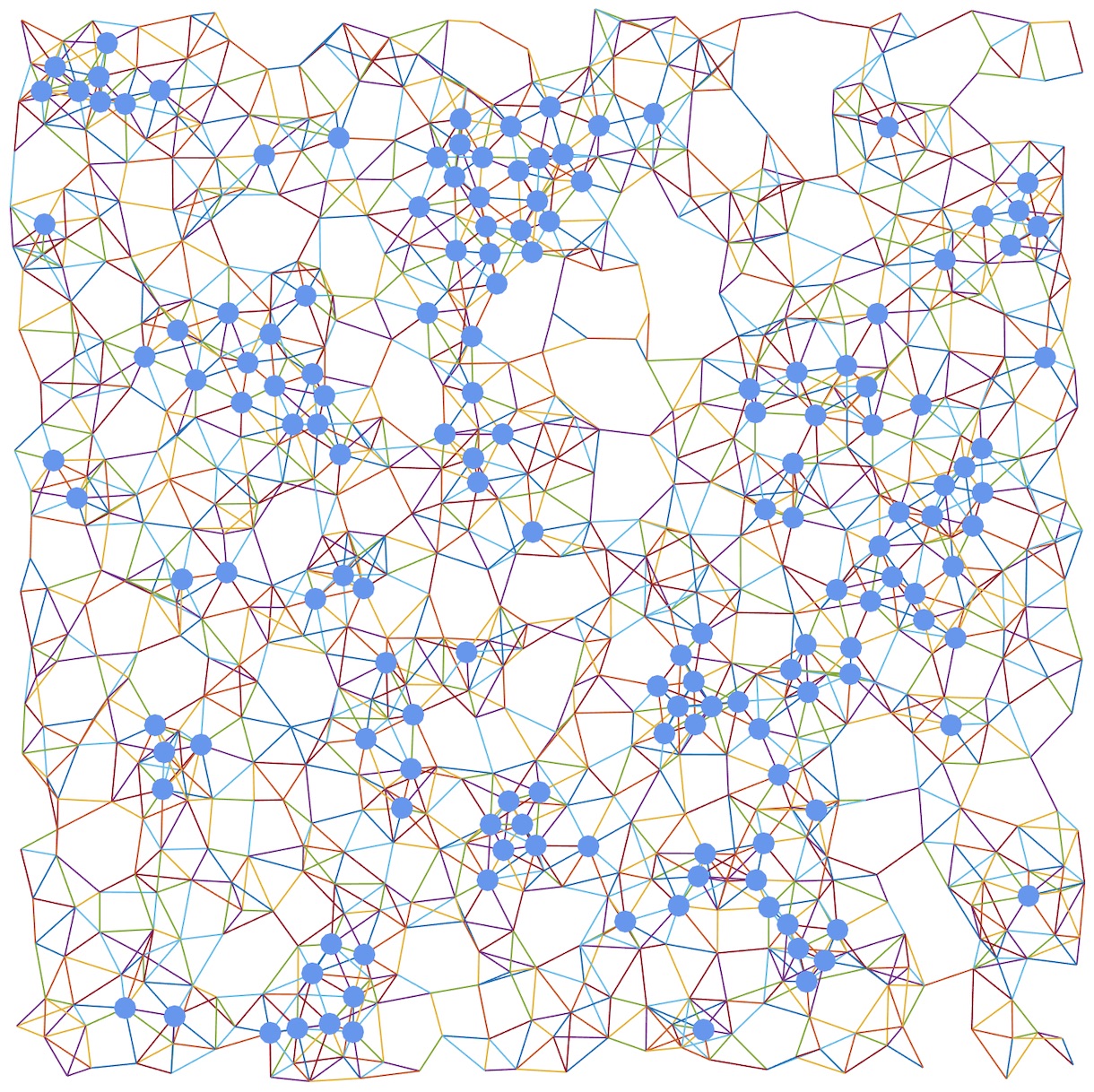}
     \end{minipage}
     \end{minipage}
    \caption{Top 20\% nodes (shown in blue) by different measures of node importance of an RSA pattern from PointPattern dataset. From left to right are results from: Degree Centrality, Eigenvector Centrality, MET matrix without normalization, MET matrix and Hybrid PANPool.}
    \label{fig:panpool_pointpattern}
    \vskip -0.1in
\end{figure}

To better understand the effect of the proposed method, in Figure~\ref{fig:panpool_pointpattern}, we visualize the top 20\% nodes by different measures of node importance of a connected point pattern called RSA, which we detail in Section~\ref{sec:pan_pointpattern}. It is noteworthy that while DC selects points relatively uniform, the result of EC is highly concentrated. This phenomenon is analogous to the contrast between the rather uniform diffusion in the classical picture and the Anderson localization \cite{anderson1958absence} in the quantum mechanics of disordered systems \cite{burda2009localization}. In this sense, it tries to find a ``mesoscopic" description that best fits the structure of input data. Importantly, we note that the unnormalized MET matrix tends to focus on the densely connected areas or hubs. In contrast, the normalized one tends to choose the \textit{locally} representative nodes and leave out the equally well-connected nodes in the hubs. This observation leads us to propose an improved pooling strategy that balances the influencers at both the global and local levels.

\paragraph{Hybrid PANPool}
To combine the contribution of the local motifs and the global importance, we propose a hybrid PAN pooling (denoted by PANPool) using a simple linear model. The global importance can be represented by, but not limited to the strength of the input signal $X$ itself. More precisely, we project feature $X \in R^{N\times d}$ by a trainable parameter vector $p\in R^d$ and combine it with the diagonal ${\rm diag}(M)$ of the MET matrix to obtain a score vector 
\begin{equation}\label{eq:score_pool}
    {\rm score} = Xp + \beta {\rm diag}(M).
\end{equation}
Here $\beta$ is a real learnable parameter that controls the emphasis on these two potentially competing factors. PANPool then selects a fraction of the nodes ranked by this score, and outputs the pooled feature array $\widetilde{X} \in R^{K\times d}$ and the corresponding adjacency matrix $\widetilde{A}\in R^{K\times K}$. This new node score in \eqref{eq:score_pool} has jointly considered both node features (at global level) and graph structures (at local level). In Figure~\ref{fig:panpool_pointpattern}, PANPool tends to select nodes that are both important locally and globally. We also tested alternative designs under the same consideration, see supplementary material for details.

\section{Related Works}
Graph neural networks have received much attention recently \cite{Survey_Battaglia,LiTaBrZe2015,scarselli2009graph,Survey_ZhangCQ,
Survey_ZhuWW,Survey_SunMS}. For graph convolutions, many works take accounts of the first order of the adjacency matrix in the spatial domain or graph Laplacian in the spectral domain. Bruna et al. \cite{BrZaSzLe2013} first proposed graph convolution using the Fourier method, which is, however, computationally expensive. Many different methods have been proposed to overcome this difficulty \cite{DCNN_2016,ChZhSo2018,ChMaXi2018fastgcn,defferrard2016convolutional,gilmer2017neural,hamilton2017inductive,KiWe2017,Monti_etal2017,Graph-CNN_2017,GWNN,GIN}.  
Another vital stream considers the attention mechanism \cite{velivckovic2017graph}, which infers the interaction between nodes without using a diffusion-like picture. Some other GNN models use multi-scale information and higher-order adjacency matrix \cite{sami2018watch,mixhop,ngcn,flam2020neural,klicpera2019diffusion,LNet,SGC}. Compared to the generic diffusion picture \cite{node2vec_2016,DeepWalk_2014,tang2015line}, the maximal entropy random walk has already shown excellent performance on link prediction \cite{li2011link} or community detection \cite{ochab2013maximal} tasks. However, many popular models can be related to or viewed as certain explicit realizations of our framework. We can interpret the MET matrix as an operator that acts on the graph input, which works as a kernel that allocates appropriate weights among the neighbors of a given node. This mechanism is similar to the attention mechanism \cite{velivckovic2017graph}, while we restrict the functional form of $M$ based on physical intuitions and preserve a compact form. Although we keep the number of features by applying $M$, one can easily concatenate the aggregated information of neighbors like GraphSAGE \cite{hamilton2017inductive} or GAT \cite{velivckovic2017graph}. Importantly, the best choice of the cutoff $L$ reveals the intrinsic dynamics of the graph. In particular, by choosing $L=1$, model \eqref{eq:conv_symm_norm} is essentially the GCN model \cite{KiWe2017}. The trick of adding self-loops is automatically realized in higher powers of $A$. By replacing $A$ in \eqref{eq:conv_symm_norm} with $D^{-1}A$ or $D^{-\frac{1}{2}}AD^{-\frac{1}{2}}$, we can easily transform our model to a multi-step GRW version, which is indeed the format of LanczosNet \cite{LNet}. The preliminary ideas about PAN convolution and its application to node classification have been presented at an ICML workshop \cite{MaLiWa2019}. This paper focuses on path integral based convolution and pooling for classification and regression tasks at graph-level.

Graph pooling is another crucial step of a GNN to make the output uniform size in graph classification and regression tasks. Researchers have proposed many pooling methods from different aspects. For example, one can merely consider node feature or node embeddings \cite{duvenaud2015convolutional, gilmer2017neural,vinyals2015order,zhang2018end}. These global pooling methods do not utilize the hierarchical structure of the graph. One way to reinforce learning ability is to build a data-dependent pooling layer with trainable operations or parameters \cite{cangea2018towards, gao2019graph, knyazev2019understanding,lee2019self,ying2018hierarchical}. One can incorporate more edge information in graph pooling \cite{diehl2019towards, Yuan2020StructPool}. One can also use spectral method and pool in Fourier or wavelet domain \cite{ma2019graph,noutahi2019towards,wang2020haargraph}.
PANPool is a method that takes both feature and structure into account.
Finally, it does not escape our analysis that the loss of paths could represent an efficient way to achieve dropout.

\section{Experiments}
In this section, we present the test results of PAN on various datasets in graph classification tasks. We show a performance comparison of PAN with some existing GNN methods. All the experiments were performed using PyTorch Geometric \cite{fey2019fast} and run on a server with Intel(R) Core(TM) i9-9820X CPU 3.30GHz, NVIDIA GeForce RTX 2080 Ti and NVIDIA TITAN V GV100. 

\subsection{PAN on Graph Classification Benchmarks}
\paragraph{Datasets and baseline methods}
We test the performance of PAN on five widely used benchmark datasets for graph classification tasks~\cite{KKMMN2016}, including two protein graph datasets \textbf{PROTEINS} and \textbf{PROTEINS\_full} ~\cite{borgwardt2005protein,dobson2003distinguishing};
one mutagen dataset \textbf{MUTAGEN}~\cite{riesen2008iam,kazius2005derivation} (full name Mutagenicity); and one dataset that consists of chemical compounds screened for activity against non-small cell lung cancer and ovarian cancer cell lines \textbf{NCI1}~\cite{wale2008comparison}; one dataset that consists of molecular compounds for activity against HIV or not \textbf{AIDS}~\cite{riesen2008iam}.
These datasets cover different domains, sample sizes, and graph structures, thus enable us to obtain a comprehensive understanding of PAN's performance in various scenarios. Specifically, the number of data samples ranges from 1,113 to 4,337, the average number of nodes is from 15.69 to 39.06, and the average number of edges is from 16.20 to 72.82, see a detailed statistical summary of the datasets in the supplementary material.
We compare \textbf{PAN} in Table~\ref{tab:pan_benchmark} with existing GNN models built by combining graph convolution layers \textbf{GCNConv} \cite{KiWe2017}, \textbf{SAGEConv} \cite{hamilton2017inductive}, \textbf{GATConv} \cite{velivckovic2017graph}, or \textbf{SGConv} \cite{Wu2019Simplifying}, and graph pooling layers \textbf{TopKPool}, \textbf{SAGPool} \cite{lee2019self}, \textbf{EdgePool} \cite{ma2019graph}, or \textbf{ASAPool} \cite{ranjan2019asap}.

\vspace{-2mm}
\paragraph{Setting}
In each experiment, we split 80\% and 20\% of each dataset for training and test. All GNNs models shared the exactly same architecture: Conv($n_f$-512) + Pool + Conv(512-256) + Pool + Conv(256-128) + FC(128-$n_c$), where $n_f$ is the feature dimension and $n_c$ is the number of classes. We give the choice of hyperparameters for these layers in the supplementary material. We evaluate the performance by the percentage of correctly predicted labels on test data. Specifically for PAN, we compared different choices of the cutoff $L$ (between 2 and 7) and reported the one that achieved the best result (shown in the brackets of Table~\ref{tab:pan_benchmark}).

\vspace{-2mm}
\paragraph{Results}
Table~\ref{tab:pan_benchmark} reports classification test accuracy for several GNN models. PAN has excellent performance on all datasets and achieves top accuracy on four of the five datasets, and in some cases, improve state of the art by a few percentage points. Even for MUTAGEN, PAN still has the second-best performance. 
Most interestingly, the optimal choice of the highest order $L$ for the MET matrix varies for different types of graph data. It confirms that the flexibility of PAN enables it to learn and adapt to the most natural representation of the given graph data.

Additionally, we also tested PAN on graph regression tasks such as QM7 and achieved excellent performances. See supplementary material for details.

\begin{table*}[t]
\centering
\begin{minipage}{\textwidth}
\centering
	\caption{Performance comparison for graph classification tasks
(test accuracy in percentage; bold font is used to highlight the best performance in the list; the value in brackets is the cutoff $L$ used in the MET matrix.)}\label{tab:pan_benchmark}
\end{minipage}
\begin{center}
\begin{small}
\begin{threeparttable}
\begin{tabularx}{390pt}{l *6{>{\Centering}X}}
\toprule
\newcommand{\nz}{\phantom{*}}
{\bf Method} &  PROTEINS & PROTEINSF & NCI1 & AIDS & MUTAGEN  \\
\midrule
GCNConv + TopKPool         
	&67.71	&68.16	&50.85	&79.25	&58.99 \\
SAGEConv + SAGPool         
	&64.13	&70.40	&64.84	&77.50	&67.40 \\
GATConv + EdgePool         
	&64.57	&62.78	&59.37	&79.00	&62.33 \\
SGConv + TopKPool      
   &68.16	&69.06	&50.85	&79.00	&63.82 \\
GATConv + ASAPool      
   &64.57	&65.47	&50.85	&79.25	&56.68 \\
SGConv + EdgePool     
	&70.85	&69.51	&56.33	&79.00	&\textbf{70.05} \\
SAGEConv + ASAPool    
	&58.74	&58.74	&50.73	&79.25	&56.68 \\
GCNConv + SAGPool    
	&59.64	&\textbf{72.65}	&50.85	&78.75	&67.28 \\
\midrule
PANConv+PANPool (ours)
	&\textbf{73.09} (1)	&\textbf{72.65} (1)	&\textbf{68.98} (3)	&\textbf{92.75} (2)	&69.70 (2)\\
\bottomrule
\end{tabularx}
\centering
\end{threeparttable}
\end{small}
\end{center}
\vskip -0.2in
\end{table*}

\subsection{PAN for Point Distribution Recognition}\label{sec:pan_pointpattern}
\paragraph{A new classification dataset for point pattern recognition}
People have proposed many graph neural network architectures; however, there are still insufficient well-accepted datasets to access their relative strength \cite{hu2020open}. Despite being popular, many datasets suffer from a lack of understanding of the underlying mechanism, such as whether one can theoretically guarantee that a graph representation is proper. These datasets are usually not controllable either; many different prepossessing tricks might be needed, such as zero paddings. Consequentially, reproducibility might be compromised.
\begin{figure}[th]
\vskip -1mm
    \centering
    \begin{minipage}{0.85\textwidth}
    \centering
    \includegraphics[width=0.25\textwidth]{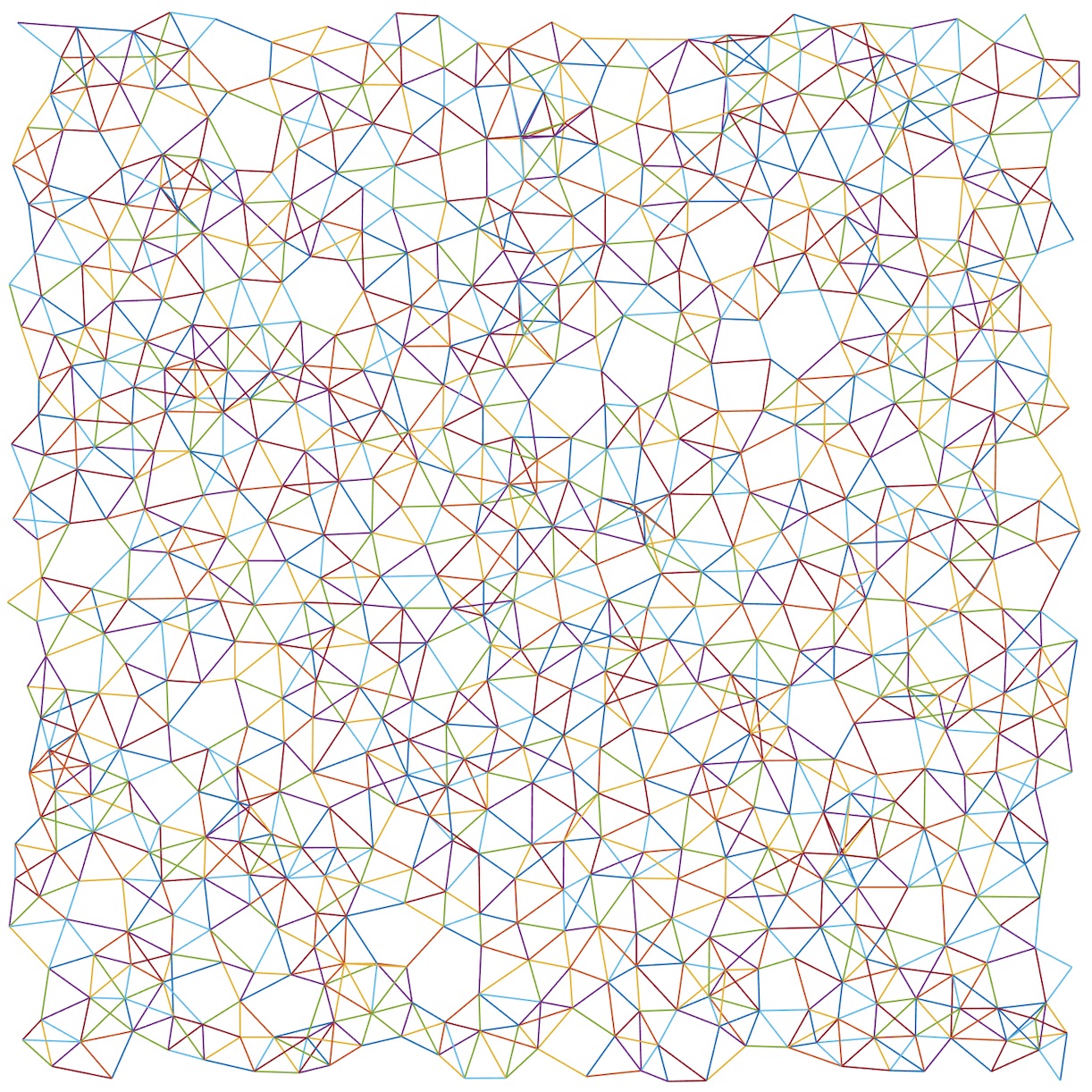}
    \hspace{2mm}
    \includegraphics[width=0.25\textwidth]{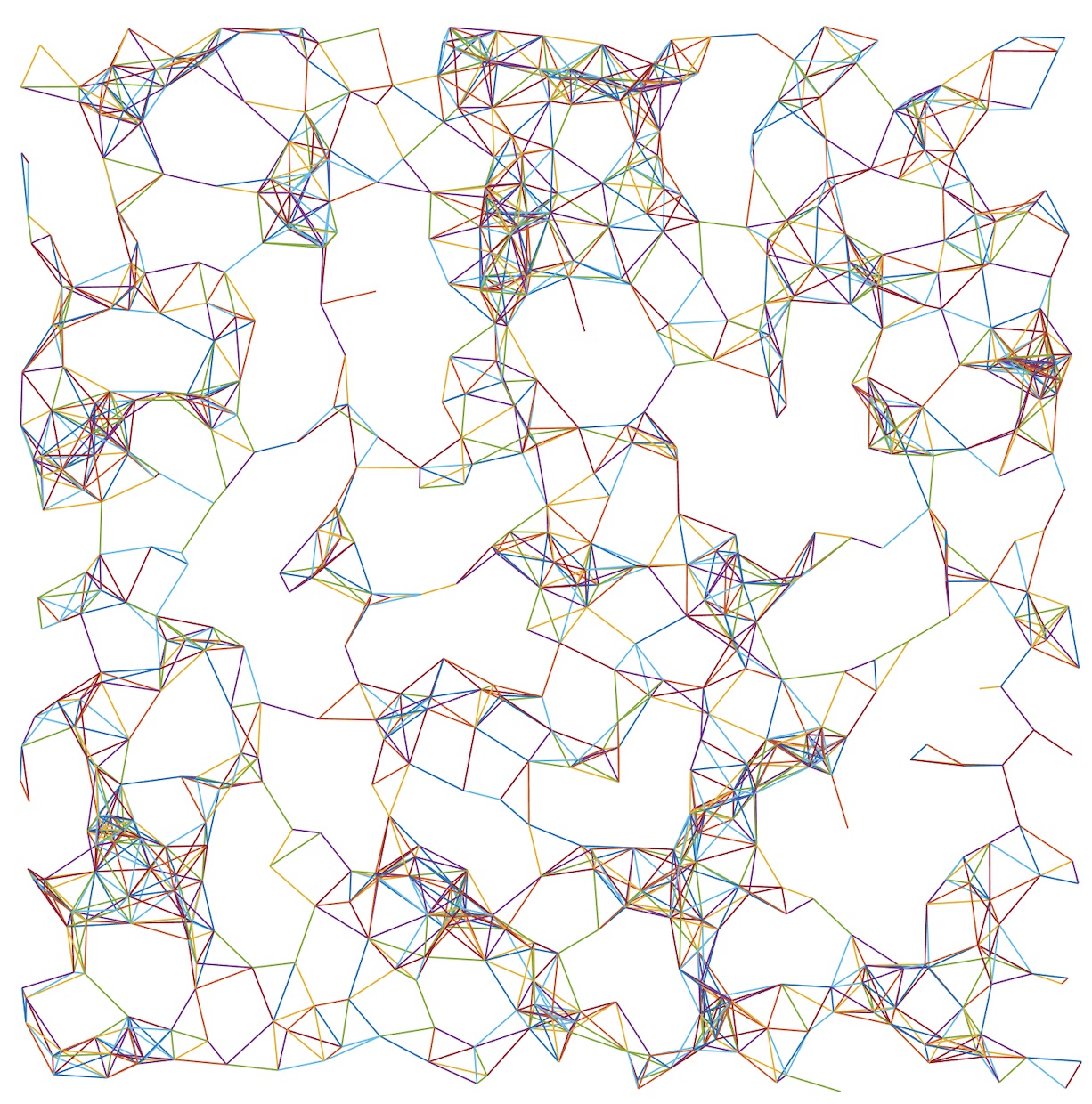}
    \hspace{2mm}
    \includegraphics[width=0.25\textwidth]{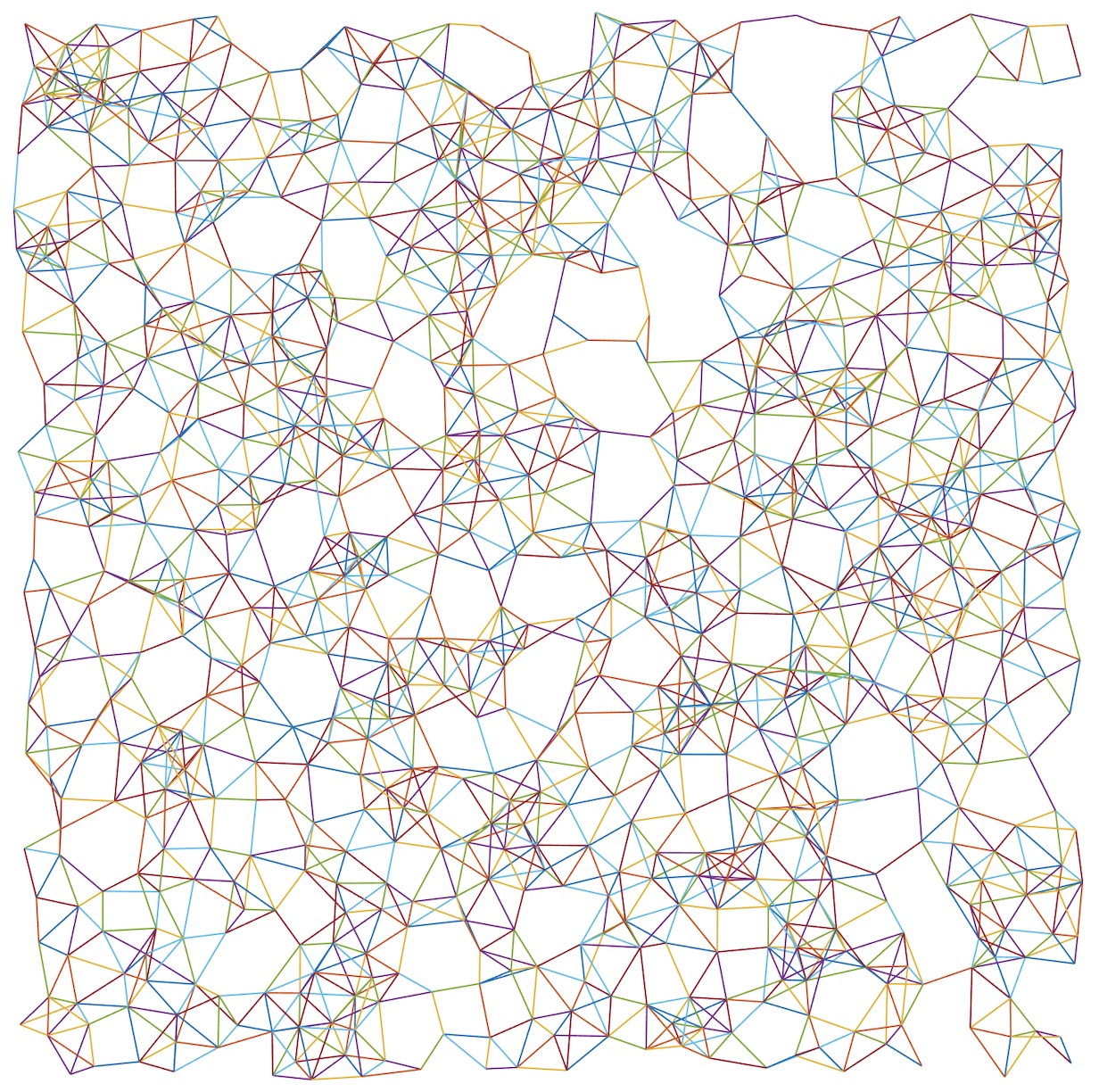}
    \end{minipage}
    \caption{From left to right: Graph samples generated from HD, Poisson and RSA point processes in PointPattern dataset.}
    \label{fig:hpr_examples}
    \vskip -2mm
\end{figure}
In order to tackle this challenge, we introduce a new graph classification dataset constructed by simple point patterns from statistical mechanics. We simulated three point patterns in 2D: hard disks in equilibrium (HD), Poisson point process, and random sequential adsorption (RSA) of disks. The HD and Poisson distributions can be seen as simple models that describe the microstructures of liquids and gases \cite{hansen1990theory}, while the RSA is a nonequilibrium stochastic process that introduces new particles one by one subject to nonoverlapping conditions. These systems are well known to be structurally different, while being easy to simulate, thus provides a solid and controllable classification task. For each point pattern, the particles are treated as nodes, and edges are subsequently drawn according to whether two particles are within a threshold distance. We name the dataset \textbf{PointPattern}. See Figure~\ref{fig:hpr_examples} for an example of the three types of resulting graphs. 
The volume fraction (covered by particles) $\phi_{\rm HD}$ of HD is fixed at 0.5,  while we tune $\phi_{\rm RSA}$ to control the similarity between RSA and the other two distributions (Poisson point pattern corresponds to $\phi_{\rm RSA}$=0). As $\phi_{\rm RSA}$ becomes closer to 0.5, RSA patterns are harder to be distinguished from HD. We use the degree as the feature for each node. It thus allows us to generate a series of graph datasets with varying difficulties as classification tasks.

\begin{figure}[th]
\vskip -1mm
    \centering
    \begin{minipage}{0.8\textwidth}
    \includegraphics[width=0.45\textwidth]{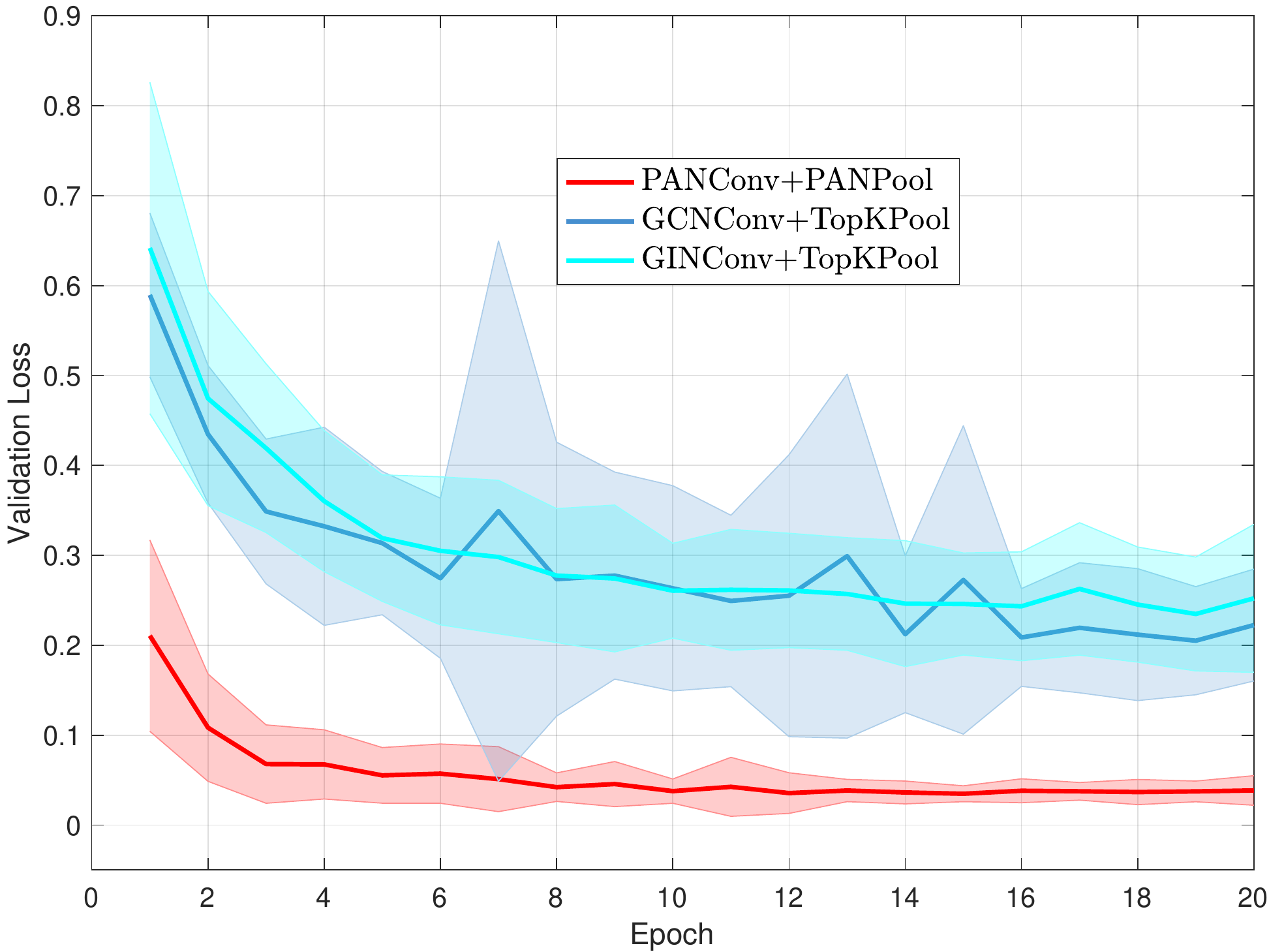}
    \hspace{3mm}
    \includegraphics[width=0.45\textwidth]{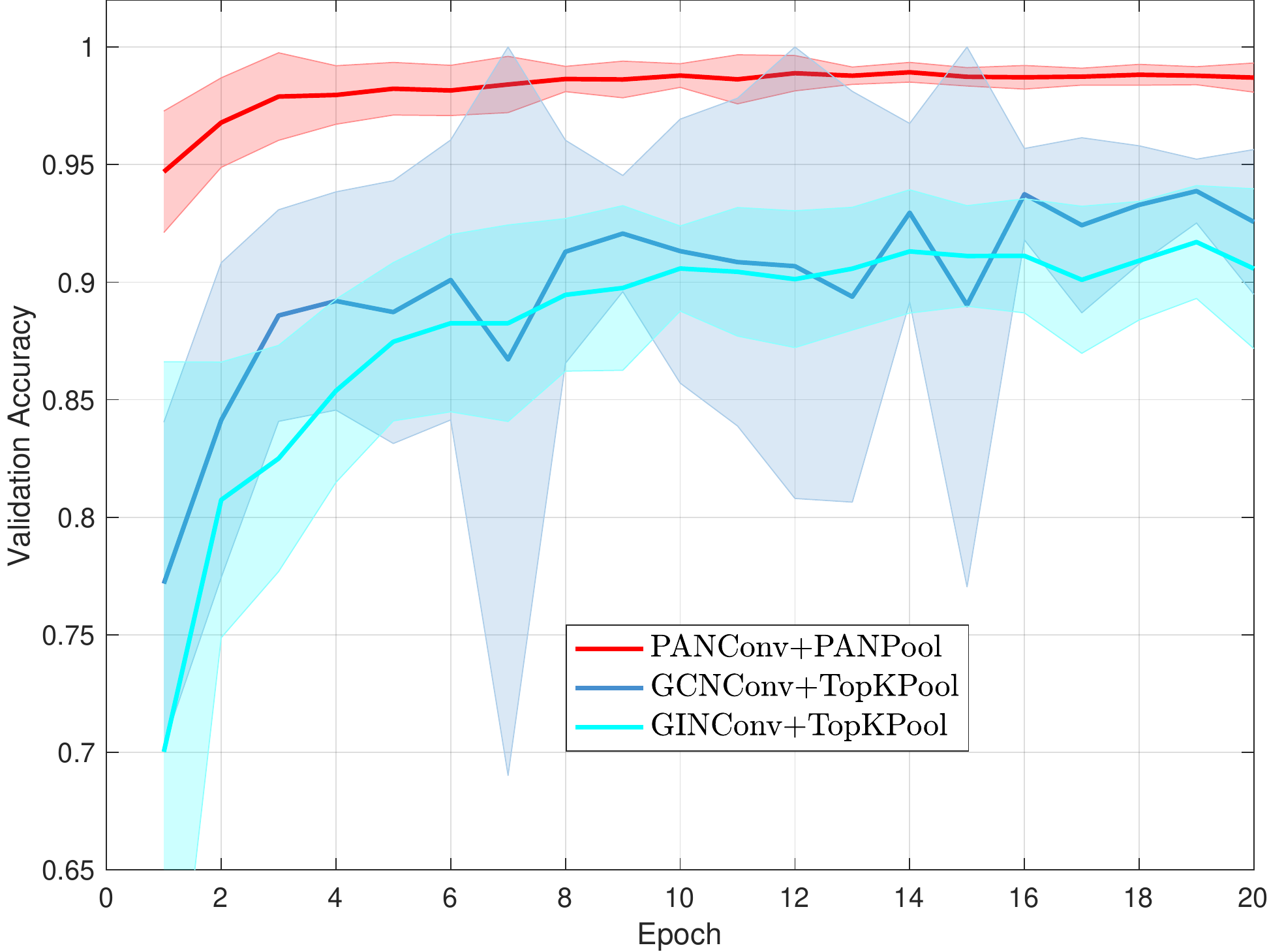}
    \end{minipage}
    \caption{Comparison of validation loss and accuracy of PAN, GCN and GIN on PointPattern under similar network architectures with 10 repetitions.}
    \label{fig:hpr_pan_gcn_gin}
    \vskip -3mm
\end{figure}

\paragraph{Setting} We tested the \textbf{PANConv+PANPool} model on \textbf{PointPattern} with $\phi_{\rm RSA}=0.3, 0.35$ and $0.4$, and compared it with other two GNN models which use \textbf{GCNConv+TopKPool} or \textbf{GINConv+TopKPool} as basic architecture blocks \cite{cangea2018towards,gao2019graph,KiWe2017,knyazev2019understanding,GIN}. Each \textbf{PointPattern} dataset is a 3-classification problem for 15,000 graphs (5000 for each type) with sizes varying between 100 and 1000.
All GNN models use the same network architecture: 3 units of one graph convolutional layer plus one graph pooling, followed by fully connected layers. In GCN and GIN models, we also use global max pooling to compress the node size to one before the fully connected layer.
We split the data into training, validation, and test sets of size 12,000, 1,500, and 1,500. We fix the number of neurons in the convolutional layers to 64, the learning rate and weight decay are set to 0.001 and 0.0005.

\begin{table}[thbp!]
\begin{minipage}{\textwidth}
	\caption{
	Test accuracy (in percentage) of PAN, GIN and GCN on three types of PointPattern datasets with different difficulties, epoch up to 20. The value in brackets is the cutoff of $L$.}\label{tab:pointpattern_pan_gcn_gin}\vspace{-2mm}
\begin{center}
\begin{small}
\begin{tabularx}{370pt}{lccc}\toprule
\textbf{PointPattern} & GINConv + SAGPool & GCNConv + TopKPool & PANConv + PANPool (ours)\\
\midrule
$\phi_{\rm RSA}=0.3$ & 90.9$\pm$2.95 & 92.9$\pm$3.21 & 99.0$\pm$0.30 (4)\\
$\phi_{\rm RSA}=0.35$ & 86.7$\pm$3.30 & 89.3$\pm$3.31 & 97.6$\pm$0.53 (4)\\
$\phi_{\rm RSA}=0.4$ & 80.2$\pm$3.80 & 85.1$\pm$4.06 & 94.4$\pm$0.55 (4)\\
\bottomrule
\end{tabularx}
\end{small}
\end{center}
\vskip -5mm
\end{minipage}
\end{table}

\vspace{-3mm}
\paragraph{Results} Table~\ref{tab:pointpattern_pan_gcn_gin} shows the mean and SD of the test accuracy of the three networks on the three PointPattern datasets. PAN outperforms GIN and GCN models on all datasets with 5 to 10 percents higher accuracy, while significantly reduces variances. We observe that PAN's advantage is persistent over varying task difficulties, which may be due to the consideration of higher order paths (here $L=4$).
We compare the validation loss and accuracy trends in the training of PANConv+PANPool with GCNConv+TopKPool in Figure~\ref{fig:hpr_pan_gcn_gin}. It illustrates that the learning and generalization capabilities of PAN are better than those of the GCN and GIN models. The loss of PAN decays to much smaller values early while the accuracy reaches higher plateau more rapidly. Moreover, the loss and accuracy of PAN both have much smaller variances, which can be seen most evidently after epoch four. In this perspective, PAN provides a more efficient and stable learning model for the graph classification task.
Another intriguing pattern we notice is that the weights are concentrated on the powers $A^3$ and $A^4$. It suggests that what differentiates these graph structures is the high orders of the adjacency matrix, or physically, the pair correlations at intermediate $r$. It may explain why PAN performs better than GCN, which uses only $A$ in its model.

\vspace{-1mm}
\section{Conclusion}
\vspace{-2mm}
We propose a path integral based GNN framework (PAN), which consists of self-consistent convolution and pooling units, the later is closely related to the subgraph centrality. PAN can be seen as a class of generalization of GNN. PAN achieves excellent performances on various graph classification and regression tasks, while demonstrating fast convergence rate and great stability. We also introduce a new graph classification dataset \textbf{PointPattern} which can serve as a new benchmark.


\bibliographystyle{plain}
\bibliography{references.bib}

\begin{thebibliography}{10}

\bibitem{sami2018watch}
Sami Abu-El-Haija, Bryan Perozzi, Rami Al-Rfou, and Alexander~A Alemi.
\newblock Watch your step: Learning node embeddings via graph attention.
\newblock In {\em NeurIPS}, pages 9180--9190, 2018.

\bibitem{mixhop}
Sami Abu-El-Haija, Bryan Perozzi, Amol Kapoor, Hrayr Harutyunyan, Nazanin
  Alipourfard, Kristina Lerman, Greg~Ver Steeg, and Aram Galstyan.
\newblock Mixhop: Higher-order graph convolution architectures via sparsified
  neighborhood mixing.
\newblock In {\em ICML}, 2019.

\bibitem{ngcn}
Sami Abu-El-Haija, Bryan Perozzi, Amol Kapoor, and Joonseok Lee.
\newblock {N-GCN:} multi-scale graph convolutionfor semi-supervised node
  classification.
\newblock In {\em UAI}, 2019.

\bibitem{AlRaPaPa2017}
Han Altae-Tran, Bharath Ramsundar, Aneesh~S Pappu, and Vijay Pande.
\newblock Low data drug discovery with one-shot learning.
\newblock {\em ACS Central Science}, 3(4):283--293, 2017.

\bibitem{anderson1958absence}
Philip~W Anderson.
\newblock Absence of diffusion in certain random lattices.
\newblock {\em Physical Review}, 109(5):1492, 1958.

\bibitem{DCNN_2016}
James Atwood and Don Towsley.
\newblock Diffusion-convolutional neural networks.
\newblock In {\em NIPS}, pages 1993--2001, 2016.


\bibitem{barabasi2016network}
Albert-L{\'a}szl{\'o} Barab{\'a}si et~al.
\newblock {\em Network Science}.
\newblock Cambridge University Press, 2016.

\bibitem{Survey_Battaglia}
Peter~W Battaglia, Jessica~B Hamrick, Victor Bapst, Alvaro Sanchez-Gonzalez,
  Vinicius Zambaldi, Mateusz Malinowski, Andrea Tacchetti, David Raposo, Adam
  Santoro, Ryan Faulkner, et~al.
\newblock Relational inductive biases, deep learning, and graph networks.
\newblock {\em arXiv preprint arXiv:1806.01261}, 2018.

\bibitem{BlRe2009}
Lorenz~C Blum and Jean-Louis Reymond.
\newblock 970 million druglike small molecules for virtual screening in the
  chemical universe database gdb-13.
\newblock {\em Journal of the American Chemical Society}, 131(25):8732--8733,
  2009.

\bibitem{borgwardt2005protein}
Karsten~M Borgwardt, Cheng~Soon Ong, Stefan Sch{\"o}nauer, SVN Vishwanathan,
  Alex~J Smola, and Hans-Peter Kriegel.
\newblock Protein function prediction via graph kernels.
\newblock {\em Bioinformatics}, 21(suppl\_1):i47--i56, 2005.

\bibitem{Breiman2001}
Leo Breiman.
\newblock Random forests.
\newblock {\em Machine Learning}, 45(1):5--32, 2001.

\bibitem{Bronstein_etal2017}
Michael~M Bronstein, Joan Bruna, Yann LeCun, Arthur Szlam, and Pierre
  Vandergheynst.
\newblock Geometric deep learning: going beyond euclidean data.
\newblock {\em IEEE Signal Processing Magazine}, 34(4):18--42, 2017.

\bibitem{BrZaSzLe2013}
Joan Bruna, Wojciech Zaremba, Arthur Szlam, and Yann LeCun.
\newblock Spectral networks and locally connected networks on graphs.
\newblock In {\em ICLR}, 2014.

\bibitem{burda2009localization}
Zdzislaw Burda, Jarek Duda, Jean-Marc Luck, and Bartek Waclaw.
\newblock Localization of the maximal entropy random walk.
\newblock {\em Physical Review Letters}, 102(16):160602, 2009.

\bibitem{cangea2018towards}
C{\u{a}}t{\u{a}}lina Cangea, Petar Veli{\v{c}}kovi{\'c}, Nikola Jovanovi{\'c},
  Thomas Kipf, and Pietro Li{\`o}.
\newblock Towards sparse hierarchical graph classifiers.
\newblock In {\em NeurIPS Workshop on Relational Representation Learning},
  2018.

\bibitem{ChZhSo2018}
Jianfei Chen, Jun Zhu, and Le~Song.
\newblock Stochastic training of graph convolutional networks with variance
  reduction.
\newblock In {\em ICML}, pages 941--949, 2018.

\bibitem{ChMaXi2018fastgcn}
Jie Chen, Tengfei Ma, and Cao Xiao.
\newblock {FastGCN: fast learning with graph convolutional networks via
  importance sampling}.
\newblock In {\em ICLR}, 2018.

\bibitem{CoVa1995}
Corinna Cortes and Vladimir Vapnik.
\newblock Support-vector networks.
\newblock {\em Machine Learning}, 20(3):273--297, 1995.

\bibitem{defferrard2016convolutional}
Micha{\"e}l Defferrard, Xavier Bresson, and Pierre Vandergheynst.
\newblock Convolutional neural networks on graphs with fast localized spectral
  filtering.
\newblock In {\em NIPS}, pages 3844--3852, 2016.

\bibitem{diehl2019towards}
Frederik Diehl, Thomas Brunner, Michael~Truong Le, and Alois Knoll.
\newblock Towards graph pooling by edge contraction.
\newblock In {\em ICML Workshop on Learning and Reasoning with Graph-Structured
  Representation}, 2019.

\bibitem{dobson2003distinguishing}
Paul~D Dobson and Andrew~J Doig.
\newblock Distinguishing enzyme structures from non-enzymes without alignments.
\newblock {\em Journal of Molecular Biology}, 330(4):771--783, 2003.

\bibitem{duvenaud2015convolutional}
David~K Duvenaud, Dougal Maclaurin, Jorge Iparraguirre, Rafael Bombarell,
  Timothy Hirzel, Al{\'a}n Aspuru-Guzik, and Ryan~P Adams.
\newblock Convolutional networks on graphs for learning molecular fingerprints.
\newblock In {\em NIPS}, pages 2224--2232, 2015.

\bibitem{estrada2005subgraph}
Ernesto Estrada and Juan~A Rodriguez-Velazquez.
\newblock Subgraph centrality in complex networks.
\newblock {\em Physical Review E}, 71(5):056103, 2005.

\bibitem{fey2019fast}
Matthias Fey and Jan~Eric Lenssen.
\newblock Fast graph representation learning with pytorch geometric.
\newblock In {\em ICLR Workshop on Representation Learning on Graphs and
  Manifolds}, 2019.

\bibitem{feynman1948space}
Richard~P Feynman.
\newblock Space-time approach to non-relativistic quantum mechanics.
\newblock {\em Reviews of Modern Physics}, 20:367--387, Apr 1948.

\bibitem{feynman2010quantum}
Richard~P Feynman, Albert~R Hibbs, and Daniel~F Styer.
\newblock {\em Quantum mechanics and path integrals}.
\newblock Courier Corporation, 2010.

\bibitem{flam2020neural}
Daniel Flam-Shepherd, Tony Wu, Pascal Friederich, and Alan Aspuru-Guzik.
\newblock Neural message passing on high order paths.
\newblock {\em arXiv preprint arXiv:2002.10413}, 2020.

\bibitem{gao2019graph}
Hongyang Gao and Shuiwang Ji.
\newblock {Graph U-Nets}.
\newblock In {\em ICML}, pages 2083--2092, 2019.

\bibitem{gilmer2017neural}
Justin Gilmer, Samuel~S Schoenholz, Patrick~F Riley, Oriol Vinyals, and
  George~E Dahl.
\newblock Neural message passing for quantum chemistry.
\newblock In {\em ICML}, pages 1263--1272, 2017.

\bibitem{node2vec_2016}
Aditya Grover and Jure Leskovec.
\newblock node2vec: Scalable feature learning for networks.
\newblock In {\em KDD}, pages 855--864, 2016.

\bibitem{hamilton2017inductive}
Will Hamilton, Zhitao Ying, and Jure Leskovec.
\newblock Inductive representation learning on large graphs.
\newblock In {\em NIPS}, pages 1024--1034, 2017.

\bibitem{hansen1990theory}
Jean-Pierre Hansen and Ian~R McDonald.
\newblock {\em Theory of simple liquids}.
\newblock Elsevier, 1990.

\bibitem{hu2020open}
Weihua Hu, Matthias Fey, Marinka Zitnik, Yuxiao Dong, Hongyu Ren, Bowen Liu,
  Michele Catasta, and Jure Leskovec.
\newblock Open graph benchmark: Datasets for machine learning on graphs.
\newblock {\em arXiv preprint arXiv:2005.00687}, 2020.

\bibitem{kazius2005derivation}
Jeroen Kazius, Ross McGuire, and Roberta Bursi.
\newblock Derivation and validation of toxicophores for mutagenicity
  prediction.
\newblock {\em Journal of Medicinal Chemistry}, 48(1):312--320, 2005.

\bibitem{KKMMN2016}
Kristian Kersting, Nils~M. Kriege, Christopher Morris, Petra Mutzel, and Marion
  Neumann.
\newblock Benchmark data sets for graph kernels, 2020.
\newblock http://www.graphlearning.io/.

\bibitem{KiWe2017}
Thomas~N Kipf and Max Welling.
\newblock Semi-supervised classification with graph convolutional networks.
\newblock In {\em ICLR}, 2017.

\bibitem{kleinert2009path}
Hagen Kleinert.
\newblock {\em Path integrals in quantum mechanics, statistics, polymer
  physics, and financial markets}.
\newblock World scientific, 2009.

\bibitem{klicpera2019diffusion}
Johannes Klicpera, Stefan Wei{\ss}enberger, and Stephan G\"{u}nnemann.
\newblock Diffusion improves graph learning.
\newblock In {\em NeurIPS}, pages 13354--13366, 2019.

\bibitem{knyazev2019understanding}
Boris Knyazev, Graham~W Taylor, and Mohamed~R Amer.
\newblock Understanding attention and generalization in graph neural networks.
\newblock In {\em NeurIPS}, 2019.

\bibitem{lee2019self}
Junhyun Lee, Inyeop Lee, and Jaewoo Kang.
\newblock Self-attention graph pooling.
\newblock In {\em ICML}, pages 3734--3743, 2019.

\bibitem{li2011link}
Rong-Hua Li, Jeffrey~Xu Yu, and Jianquan Liu.
\newblock Link prediction: the power of maximal entropy random walk.
\newblock In {\em CIKM}, pages 1147--1156, 2011.

\bibitem{LiTaBrZe2015}
Yujia Li, Daniel Tarlow, Marc Brockschmidt, and Richard Zemel.
\newblock Gated graph sequence neural networks.
\newblock {\em ICLR}, 2016.

\bibitem{LNet}
Renjie Liao, Zhizhen Zhao, Raquel Urtasun, and Richard~S Zemel.
\newblock Lanczosnet: Multi-scale deep graph convolutional networks.
\newblock In {\em ICLR}, 2019.

\bibitem{ma2019graph}
Yao Ma, Suhang Wang, Charu~C. Aggarwal, and Jiliang Tang.
\newblock Graph convolutional networks with {EigenPooling}.
\newblock In {\em KDD}, pages 723--731, 2019.

\bibitem{MaLiWa2019}
Zheng Ma, Ming Li, and Yu~Guang Wang.
\newblock {PAN}: Path integral based convolution for deep graph neural
  networks.
\newblock In {\em ICML Workshop on Learning and Reasoning with Graph-Structured
  Representation}, 2019.

\bibitem{Monti_etal2017}
Federico Monti, Davide Boscaini, Jonathan Masci, Emanuele Rodola, Jan Svoboda,
  and Michael~M Bronstein.
\newblock {Geometric deep learning on graphs and manifolds using mixture model
  CNNs}.
\newblock In {\em CVPR}, pages 5425--5434, 2017.

\bibitem{newman2018networks}
Mark Newman.
\newblock {\em Networks}.
\newblock Oxford university press, 2018.

\bibitem{noutahi2019towards}
Emmanuel Noutahi, Dominique Beani, Julien Horwood, and Prudencio Tossou.
\newblock Towards interpretable sparse graph representation learning with
  {Laplacian} pooling.
\newblock {\em arXiv preprint arXiv:1905.11577}, 2019.

\bibitem{ochab2013maximal}
JK~Ochab and Zdzis{\l}aw Burda.
\newblock Maximal entropy random walk in community detection.
\newblock {\em The European Physical Journal Special Topics}, 216(1):73--81,
  2013.

\bibitem{DeepWalk_2014}
Bryan Perozzi, Rami Al-Rfou, and Steven Skiena.
\newblock Deepwalk: Online learning of social representations.
\newblock In {\em KDD}, pages 701--710, 2014.

\bibitem{Ramsundar_etal2015}
Bharath Ramsundar, Steven Kearnes, Patrick Riley, Dale Webster, David
  Konerding, and Vijay Pande.
\newblock Massively multitask networks for drug discovery.
\newblock {\em arXiv preprint arXiv:1502.02072}, 2015.

\bibitem{ranjan2019asap}
Ekagra Ranjan, Soumya Sanyal, and Partha~Pratim Talukdar.
\newblock {ASAP}: Adaptive structure aware pooling for learning hierarchical
  graph representations.
\newblock {\em AAAI}, 2020.

\bibitem{riesen2008iam}
Kaspar Riesen and Horst Bunke.
\newblock {IAM} graph database repository for graph based pattern recognition
  and machine learning.
\newblock In {\em Joint IAPR International Workshops on Statistical Techniques
  in Pattern Recognition (SPR) and Structural and Syntactic Pattern Recognition
  (SSPR)}, pages 287--297. Springer, 2008.

\bibitem{RuTkMuLi2012}
Matthias Rupp, Alexandre Tkatchenko, Klaus-Robert M{\"u}ller, and O~Anatole
  Von~Lilienfeld.
\newblock Fast and accurate modeling of molecular atomization energies with
  machine learning.
\newblock {\em Physical Review Letters}, 108(5):058301, 2012.

\bibitem{scarselli2009graph}
Franco Scarselli, Marco Gori, Ah~Chung Tsoi, Markus Hagenbuchner, and Gabriele
  Monfardini.
\newblock The graph neural network model.
\newblock {\em IEEE Transactions on Neural Networks}, 20(1):61--80, 2009.

\bibitem{Graph-CNN_2017}
Felipe~Petroski Such, Shagan Sah, Miguel~Alexander Dominguez, Suhas Pillai,
  Chao Zhang, Andrew Michael, Nathan~D Cahill, and Raymond Ptucha.
\newblock Robust spatial filtering with graph convolutional neural networks.
\newblock {\em IEEE Journal of Selected Topics in Signal Processing},
  11(6):884--896, 2017.

\bibitem{tang2015line}
Jian Tang, Meng Qu, Mingzhe Wang, Ming Zhang, Jun Yan, and Qiaozhu Mei.
\newblock Line: Large-scale information network embedding.
\newblock In {\em WWW}, pages 1067--1077, 2015.

\bibitem{velivckovic2017graph}
Petar Veli{\v{c}}kovi{\'c}, Guillem Cucurull, Arantxa Casanova, Adriana Romero,
  Pietro Lio, and Yoshua Bengio.
\newblock Graph attention networks.
\newblock In {\em ICLR}, 2018.

\bibitem{vinyals2015order}
Oriol Vinyals, Samy Bengio, and Manjunath Kudlur.
\newblock Order matters: Sequence to sequence for sets.
\newblock In {\em ICLR}, 2015.

\bibitem{wale2008comparison}
Nikil Wale, Ian~A Watson, and George Karypis.
\newblock Comparison of descriptor spaces for chemical compound retrieval and
  classification.
\newblock {\em Knowledge and Information Systems}, 14(3):347--375, 2008.

\bibitem{wang2020haargraph}
Yu~Guang Wang, Ming Li, Zheng Ma, Guido Montufar, Xiaosheng Zhuang, and Yanan
  Fan.
\newblock Haar graph pooling.
\newblock In {\em ICML}, 2020.

\bibitem{Wu2019Simplifying}
Felix Wu, Amauri Souza, Tianyi Zhang, Christopher Fifty, Tao Yu, and Kilian
  Weinberger.
\newblock Simplifying graph convolutional networks.
\newblock In {\em ICML}, pages 6861--6871, 2019.

\bibitem{SGC}
Felix Wu, Tianyi Zhang, Amauri Holanda~de Souza~Jr, Christopher Fifty, Tao Yu,
  and Kilian~Q Weinberger.
\newblock Simplifying graph convolutional networks.
\newblock In {\em ICML}, 2019.

\bibitem{Wu_etal2018}
Zhenqin Wu, Bharath Ramsundar, Evan~N Feinberg, Joseph Gomes, Caleb Geniesse,
  Aneesh~S Pappu, Karl Leswing, and Vijay Pande.
\newblock {MoleculeNet: a benchmark for molecular machine learning}.
\newblock {\em Chemical Science}, 9(2):513--530, 2018.

\bibitem{Survey_ZhangCQ}
Zonghan Wu, Shirui Pan, Fengwen Chen, Guodong Long, Chengqi Zhang, and Philip~S
  Yu.
\newblock A comprehensive survey on graph neural networks.
\newblock {\em IEEE Transactions on Neural Networks and Learning Systems}.

\bibitem{GWNN}
Bingbing Xu, Huawei Shen, Qi~Cao, Yunqi Qiu, and Xueqi Cheng.
\newblock Graph wavelet neural network.
\newblock In {\em ICLR}, 2019.

\bibitem{GIN}
Keyulu Xu, Weihua Hu, Jure Leskovec, and Stefanie Jegelka.
\newblock How powerful are graph neural networks?
\newblock In {\em ICLR}, 2019.

\bibitem{Planetoid_2016}
Zhilin Yang, William~W Cohen, and Ruslan Salakhutdinov.
\newblock Revisiting semi-supervised learning with graph embeddings.
\newblock In {\em ICML}, 2016.

\bibitem{ying2018hierarchical}
Zhitao Ying, Jiaxuan You, Christopher Morris, Xiang Ren, Will Hamilton, and
  Jure Leskovec.
\newblock Hierarchical graph representation learning with differentiable
  pooling.
\newblock In {\em NeurIPS}, pages 4800--4810, 2018.

\bibitem{Yuan2020StructPool}
Hao Yuan and Shuiwang Ji.
\newblock Structpool: Structured graph pooling via conditional random fields.
\newblock In {\em ICLR}, 2020.

\bibitem{zhang2018end}
Muhan Zhang, Zhicheng Cui, Marion Neumann, and Yixin Chen.
\newblock An end-to-end deep learning architecture for graph classification.
\newblock In {\em AAAI}, 2018.

\bibitem{Survey_ZhuWW}
Ziwei Zhang, Peng Cui, and Wenwu Zhu.
\newblock Deep learning on graphs: A survey.
\newblock {\em IEEE Transactions on Knowledge and Data Engineering}, 2020.

\bibitem{Survey_SunMS}
Jie Zhou, Ganqu Cui, Zhengyan Zhang, Cheng Yang, Zhiyuan Liu, and Maosong Sun.
\newblock Graph neural networks: A review of methods and applications.
\newblock {\em arXiv preprint arXiv:1812.08434}, 2018.

\bibitem{Zinn-Justin:2009}
Jean Zinn-Justin.
\newblock {P}ath integral.
\newblock {\em Scholarpedia}, 4(2):8674, 2009.

\end{thebibliography}

\newpage
\appendix
\section{Variations of PANPool}

In the main text, we discussed the relation between the diagonal of the MET matrix and subgraph centrality, as well as the idea of combining structural information and signals to develop pooling methods. We study several alternatives of the Hybrid PANPool proposed in the paper and report experimental results on benchmark datasets.

First, we consider the subgraph centrality's direct counterpart under the PAN framework, i.e., the weighted sum of powers of $A$. Formally, we consider the score as the diagonal of the MET matrix before normalization, it writes as 
\begin{equation}\label{eq:panumpool}
     {\rm score} = {\rm diag}(Z^{\frac{1}{2}}MZ^{\frac{1}{2}}).
 \end{equation}
 
 Similarly, we can also combine this unnormalized MET matrix with projected features, i.e.,
 \begin{equation}\label{eq:panxumpool}
     {\rm score} = Xp + \beta {\rm diag}(Z^{\frac{1}{2}}MZ^{\frac{1}{2}}).
 \end{equation}
This method also considers both graph structures and signals, while the measure of structural importance is at a global rather than local level. 

We can also take simple approaches to mix structural information with signals. Most straightforwardly, we can employ the readily calculated convoluted feature $MX$ to define the score. For example, the $\ell_2$-norm of each row of $MX$ can define a score vector. The score for node $i$ can be written as
 \begin{equation}\label{eq:panmpool}
     \mathrm{score} = ||(MX)_i||_2.
 \end{equation}
 
Finally, instead of using a parameterized linear combination of the MET matrix and projected signals, we can apply the Hadamard product of the two contributions. The score then becomes 
 \begin{equation}\label{eq:panxhmpool}
     {\rm score} = Xp \circ {\rm diag}(M).
 \end{equation}
 
We use PANUMPool, PANXUMPool, PANMPool, and PANXHMPool to denote these variations of PANPool corresponding to \eqref{eq:panumpool}--\eqref{eq:panxhmpool} in the following experimental results.

\section{Datasets and extended experiments}
We put the PyTorch codes for experiments in the folder ``codes'' with dataset downloading and program execution instructions in ``README.md''.

\subsection{PointPatterns}
All simulations are performed in square simulation boxes with periodic boundary conditions. For hard disks, we use corresponding RSA configurations as initial conditions. We then perform an average of 10,000 Monte Carlo steps per particle to equilibrate the system. In the following step of converting a point pattern to a graph, we do not consider the images of the simulation boxes; that is, we do not connect particles across the boundaries. The choice of the threshold is inevitably subjective. Here we use $4R$ as the threshold, where $R$ is the radius of the corresponding hard disks with the same number density at volume fraction 0.5. This threshold is of the same order of the typical distance between two neighboring particles, which guarantees that the resulting graph is connected.    

We list the summary statistics of the three datasets of PointPattern used in the main text with $\phi_{\rm RSA}=0.3,0.35, 0.4$ in Table~\ref{tab:statistics_pointpattern}. They can be downloaded from Google Drive at

\url{https://drive.google.com/file/d/1C3ciJsteqsKFVGF8JI8-KnXhe4zY41Ss/view?usp=sharing}\\
\url{https://drive.google.com/file/d/1rsTh09FzGxHculBVrYyl5tOHD9mxqc0G/view?usp=sharing}\\
\url{https://drive.google.com/file/d/16pI974P8WzanBUPrMHIaGfeSLoksviBk/view?usp=sharing}

We also show an example in README.md of running PAN on PointPattern, which program includes downloading and preprocessing PointPattern datasets.

\begin{table*}[th]
\caption{Summary information of PointPattern datasets.}\label{tab:statistics_pointpattern}
\begin{center}
\begin{small}
\begin{tabularx}{380pt}{l *7{>{\Centering}X}}
\toprule
{\bf PointPattern}  & $\phi_{\rm RSA}=0.3$ & $\phi_{\rm RSA}=0.35$& $\phi_{\rm RSA}=0.4$ \\
\midrule
\#classes & 3& 3 &3\\
\#graphs    &15,000 & 15,000  & 15,000\\
max \#nodes  &1000 &1000  &1000 \\
min \#nodes  &100 &100  & 100 \\
avg \#nodes &478 & 474  & 475 \\
avg \#edges &3265 &3223 &3220 \\
\bottomrule
\end{tabularx}
\end{small}
\end{center}
\end{table*}

\subsection{PAN on Classification Benchmarks}
\paragraph{Extended experiments on Classification Benchmark}
We list the summary statistics of benchmark graph classification datasets in Table~\ref{tab:statistics}. In Table~\ref{tab:pan_benchmark_1}, we report the classification test accuracy for variations of PAN compared with other methods. All networks utilize the same architecture. The PAN model, in general, has excellent performance on all datasets. The table shows the variations of PAN models can achieve the state of the art performance on a variety of graph classification tasks, and in some cases, improve state of the art by a few percentage points. In particular, PANConv+PANPool tends to perform better than other methods or variations on average, as presented in the main text. While among alternative PAN pooling methods, PANConv+PANMPool tends to have the least SD.

\begin{table*}[th]
\caption{Summary statistics of benchmark graph classification datasets.}\label{tab:statistics}
\begin{center}
\begin{small}
\begin{tabularx}{380pt}{l *7{>{\Centering}X}}
\toprule
{\bf Dataset}  & MUTAG & PROTEINS & PROTEINSF & NCI1 & AIDS & MUTAGEN \\
\midrule
max \#nodes  &28 &620 &620  &111 & 95 &417\\
min \#nodes  &10 &4 &4  & 3& 2&4 \\
avg \#nodes &17.93 & 39.06 & 39.06  & 29.87 & 15.69 & 30.32\\
\# node attributes & -& 1 & 29 & - & 4 &-\\
avg \#edges &19.79 &72.82 &72.82 &32.30 & 16.20 &30.77\\
\#graphs    &188 & 1,113 & 1,113  & 4,110& 2,000 & 4,337\\
\#classes & 2& 2 & 2 &2 & 2& 2\\
\bottomrule
\end{tabularx}
\end{small}
\end{center}
\end{table*}

\begin{table*}[t]
\centering
\begin{minipage}{\textwidth}
\centering
	\caption{Performance comparison for graph classification tasks
(test accuracy in percentage; bold font is used to highlight the best performance in the list; the $L$ of all PAN-models on five datasets are $\{3, 1, 3, 3, 3\}$, respectively).}\label{tab:pan_benchmark_1}
\end{minipage}
\begin{center}
\begin{small}
\begin{threeparttable}
\begin{tabularx}{402pt}{c cccccc}
\toprule
\newcommand{\nz}{\phantom{*}}
{\bf Method} &  PROTEINS & PROTEINSF & NCI1 & AIDS & MUTAGEN  \\
\midrule
GCNConv + TopKPool         
	&64.0$\pm$0.40	&69.6$\pm$6.03	&49.9$\pm$0.50	&81.2$\pm$1.00	&63.5$\pm$6.69 \\
SAGEConv + SAGPool         
	&70.5$\pm$3.95	&63.0$\pm$2.34	&64.0$\pm$3.61	&79.5$\pm$2.02	&67.6$\pm$3.24 \\
GATConv + EdgePool         
	&72.4$\pm$1.46		&71.3$\pm$3.16		&60.1$\pm$1.76		&80.5$\pm$0.72		&\textbf{71.5$\pm$1.09} \\
SGConv + TopKPooling      
   &73.6$\pm$1.70	&65.9$\pm$1.25	&\textbf{61.5$\pm$5.11}	&81.0$\pm$0.01	&66.3$\pm$2.08 \\
GATConv + ASAPooling      
   &64.8$\pm$5.43	&67.3$\pm$4.37	&53.9$\pm$4.11	&84.7 $\pm$6.21	&58.4$\pm$5.19 \\
SGConv + EdgePooling     
	&69.0$\pm$1.74	&70.5$\pm$2.48	&58.4$\pm$1.96	&76.7$\pm$1.12	&70.7$\pm$0.69 \\
SAGEConv + ASAPooling    
	&59.2$\pm$5.84	&63.9$\pm$2.44	&53.5$\pm$2.91	&80.6$\pm$6.39	&63.1$\pm$3.74 \\
GCNConv + SAGPooling     
	&71.5$\pm$2.72	&68.6$\pm$2.25	&52.2$\pm$8.87	&83.1$\pm$1.10	&68.9$\pm$5.80 \\
\midrule
PANConv+PANUMPool (Eq~\ref{eq:panumpool})  
&67.8$\pm$0.82	&69.1$\pm$1.21	&59.2$\pm$0.69	&82.7$\pm$7.82	&70.0$\pm$2.11\\
PANConv+PANXUMPool (Eq~\ref{eq:panxumpool})  
&69.7$\pm$1.60	&\textbf{72.6$\pm$3.20}	&60.1$\pm$1.74	&86.9$\pm$3.64	&69.4$\pm$1.08\\
PANConv+PANMPool (Eq~\ref{eq:panmpool})  
&66.8$\pm$0.78	&71.0$\pm$0.60	&51.9$\pm$1.39	&80.6$\pm$0.44	&68.4$\pm$1.01 \\
PANConv+PANXHMPool (Eq~\ref{eq:panxhmpool}) 
&68.8$\pm$5.23	&69.7$\pm$1.97	&55.9$\pm$1.81	&91.4$\pm$3.39	&70.2$\pm$1.08 \\
PANConv+PANPool
	&\textbf{76.6$\pm$2.06}	&71.7$\pm$6.05	&60.8$\pm$ 3.45	&\textbf{97.5$\pm$1.86}	&70.9$\pm$2.76\\
\bottomrule
\end{tabularx}
\centering
\end{threeparttable}
\end{small}
\end{center}
\end{table*}

\subsection{Quantum Chemistry Regression}
\paragraph{QM7} In this section, we test the performance of the PAN model on the QM7 dataset. The QM7 has been utilized to measure the efficacy of machine-learning methods for quantum chemistry \cite{BlRe2009, RuTkMuLi2012}. The dataset contains 7,165 molecules, each represented by the Coulomb (energy) matrix, and labeled with the value of atomization energy. The molecules have varying node size and structure with up to 23 atoms. We view each molecule as a weighted graph: atoms are nodes, and the Coulomb matrix of the molecule is the adjacency matrix. Since the node (atom) itself does not have feature information, we set the node feature to a constant vector with components all one, so that features here are uninformative, and the learning is mainly concerned with identifying the molecule structure. The task is to predict the atomization energy value of each molecule graph, which boils down to a standard graph regression problem.

\begin{table}[ht]
\caption{Test mean absolute error (MAE) comparison on QM7, with the standard deviation over ten repetitions of the experiments. The value in brackets is the cutoff $L$.}\label{tab:qm7_results}
\begin{minipage}{\textwidth}
\begin{center}
\begin{small}
\begin{tabular}[380pt]{cl}
\toprule
Method & Test MAE\\
\midrule
Multitask \cite{Ramsundar_etal2015}  & 123.7$\pm$15.6$^{*}$ \\[1mm]
RF \cite{Breiman2001} & 122.7$\pm$4.2$^{*}$\\[1mm]
KRR \cite{CoVa1995} &  110.3$\pm$4.7$^{*}$\\[1mm]
GC \cite{AlRaPaPa2017} & \hspace{1.5mm}77.9$\pm$2.1$^{*}$\\[1mm]
GCNConv+TopKPool & \hspace{1.5mm}43.6$\pm$0.98\\[1mm]
PANConv+PANUMPool (Eq~\ref{eq:panumpool})  & \hspace{1.5mm}43.5$\pm$0.86 (1)\\[1mm]
PANConv+PANXUMPool (Eq~\ref{eq:panxumpool})  & \hspace{1.5mm}43.3$\pm$1.32 (2)\\[1mm]
PANConv+PANMPool (Eq~\ref{eq:panmpool})  & \hspace{1.5mm}43.6$\pm$0.84 (2)\\[1mm]
PANConv+PANXHMPool (Eq~\ref{eq:panxhmpool}) & \hspace{1.5mm}43.0$\pm$1.27 (1)\\[1mm]
\textbf{PANConv+PANPool} & \hspace{1.5mm}\textbf{42.8$\pm$0.63} (1) \\
\bottomrule
\end{tabular}
\begin{tablenotes}
        \item[] {\rm `*' indicates records retrieved from \cite{Wu_etal2018}, and bold font is used to highlight the best performance in the list.}
\end{tablenotes}
\end{small}
\end{center}
\end{minipage}
\end{table}

\paragraph{Experimental setting} In the experiment, we normalize the label value by subtracting the mean and scaling SD to 1. We then need to convert the predicted output to the original label domain (by re-scaling and adding the mean back). Following \cite{gilmer2017neural}, we use mean squared error (MSE) as the loss for training and mean absolute error (MAE) as the evaluation metric for validation and test.
We use the splitting percentages of 80\%, 10\%, and 10\% for training, validation, and testing. We set the hidden dimension of the PANConv and GCN layers as 64, the learning rate 5.0e-4 for Adam optimization, and the maximal epoch 50 with no early stop.
For better comparison, we repeat all experiments ten times with randomly shuffled datasets of different random seeds.

\paragraph{Comparison methods and results} We test and compare the performance (test MAE and validation loss) of PAN against the GNN model with \textbf{GCNConv+SAGPool} \cite{KiWe2017,lee2019self} and other methods including Multitask Networks (\textbf{Multitask}) \cite{Ramsundar_etal2015}, Random Forest (\textbf{RF}) \cite{Breiman2001}, Kernel Ridge Regression (\textbf{KRR}) \cite{CoVa1995}, Graph Convolutional models (\textbf{GC}) \cite{AlRaPaPa2017}.
In our test, each PAN model contains one PANConv layer plus one PAN pooling layer, followed by two fully connected layers. The GCN model has two units of GCNConv+SAGPool, followed by GCNConv plus global max pooling and one fully connected layer.
For other methods, we use their public results from \cite{Wu_etal2018} on QM7.
In Table~\ref{tab:qm7_results}, we evaluate five PAN models on QM7 compared to other methods. The PAN models achieve top average test MAE and a smaller SD than other methods.

\end{document}